\documentclass[sigconf]{acmart}
\copyrightyear{2023} 
\acmYear{2023} 
\setcopyright{acmlicensed}\acmConference[CIKM '23]{Proceedings of the 32nd ACM International Conference on Information and Knowledge Management}{October 21--25, 2023}{Birmingham, United Kingdom}
\acmBooktitle{Proceedings of the 32nd ACM International Conference on Information and Knowledge Management (CIKM '23), October 21--25, 2023, Birmingham, United Kingdom}
\acmPrice{15.00}
\acmDOI{10.1145/3583780.3615071}
\acmISBN{979-8-4007-0124-5/23/10}
\AtBeginDocument{%
  \providecommand\BibTeX{{%
    \normalfont B\kern-0.5em{\scshape i\kern-0.25em b}\kern-0.8em\TeX}}}

\acmConference[CIKM'23]{CIKM'23: ACM International Conference on Information and Knowledge Management}{October 21-- October 25, 2023}{Birmingham, UK}

\usepackage{subcaption}
\usepackage{diagbox}
\usepackage{multirow}
\usepackage{multicol}
\usepackage{algorithm}
\usepackage{enumitem}
\usepackage{algorithmic}
\usepackage{array}
\newcolumntype{A}{>{\centering}p{0.09\textwidth}}
\newcolumntype{B}{p{0.08\textwidth}}
\newcolumntype{C}{>{\centering\arraybackslash}p{0.1\textwidth}}

\begin{document}

\title{Tackling Diverse Minorities in Imbalanced Classification}

% {\color{red}{Hi Henry, I create a main.corrected.tex file to include all the changes regarding to grammars and expressions. Please use https://www.diffchecker.com/diff to find the differences.

\author{Kwei-Herng Lai \\ Daochen Zha}
\email{khlai@rice.edu}
\affiliation{%
	\institution{Rice University }
	\city{Houston, TX}
	\country{USA}}
\author{Huiyuan Chen \\ Mangesh Bendre  }
\email{hchen@visa.com}
\affiliation{%
	\institution{Visa Research}
	\city{Palo Alto, CA}
	\country{USA}}

\author{Yuzhong Chen \\ Mahashweta Das\\ Hao Yang}
\email{yuzchen@visa.com}
\affiliation{%
	\institution{Visa Research}
	\city{Palo Alto, CA}
	\country{USA}}

 \author{Xia Hu}
\email{xia.hu@rice.edu}
\affiliation{%
	\institution{Rice University}
	\city{Houston, TX}
	\country{USA}}

\renewcommand{\shortauthors}{Kwei-Herng et al.}
\begin{abstract}
% 0.3 page TODO: Fine Tune it !!!
Imbalanced datasets are commonly observed in various real-world applications, presenting significant challenges in training classifiers. When working with large datasets, the imbalanced issue can be further exacerbated, making it exceptionally difficult to train classifiers effectively. To address the problem, over-sampling techniques have been developed to linearly interpolating data instances between minorities and their neighbors. However, in many real-world scenarios such as anomaly detection, minority instances are often dispersed diversely in the feature space rather than clustered together. Inspired by domain-agnostic data mix-up, we propose generating synthetic samples iteratively by mixing data samples from both minority and majority classes. It is non-trivial to develop such a framework, the challenges include source sample selection, mix-up strategy selection, and the coordination between the underlying model and mix-up strategies. To tackle these challenges, we formulate the problem of iterative data mix-up as a Markov decision process (MDP) that maps data attributes onto an augmentation strategy. To solve the MDP, we employ an actor-critic framework to adapt the discrete-continuous decision space. This framework is utilized to train a data augmentation policy and design a reward signal that explores classifier uncertainty and encourages performance improvement, irrespective of the classifier's convergence. We demonstrate the effectiveness of our proposed framework through extensive experiments conducted on seven publicly available benchmark datasets using three different types of classifiers. The results of these experiments showcase the potential and promise of our framework in addressing imbalanced datasets with diverse minorities.

\end{abstract}




\maketitle

% \pagestyle{plain}
% \vspace{-7.5pt}
\section{Introduction}
% There are two common strategies to tackle the problem, model-centric approaches that directly develop label exploitation strategy based on unsupervised models or data-centric approaches that augment label information from limited labels to create beneficial synthetic data instances. In contrast to the model-centric solutions that heavily rely on the strong assumption on data distribution, data-centric solution is arguably more flexible as it does not make any assumption on the model so that it is generally applicable to various models~\cite{}.

% 1.5 pages TODO: Fine Tune it !!!
%  Web search and mining

% Unsupervised --> no label, high false positive rate --> label is needed

% 1. Standard AD no labels. Not effective
% 2. Supervised AD
% 3. Limited, Anomalies have diverse distributions. So it is hard
% 4. In parallel, DA has been applied XXX. Research question: Can we apple DA in Supervised AD
% 5. Challenges. 1. Diverse distribution. 2. Anomaly label imbalance. 3. Easily generate some noisy data. Toy example.
% 6. Our framework

% Real-world data are imbalanced, for example: anomalies 
Imbalanced data with a disproportionate ratio of training samples between classes widely exists in many real-world datasets and their corresponding applications, such as intrusion detection from server log~\cite{kim2020ai}, fault detection in manufacturing~\cite{hsu2021multiple}, and fraud detection in finance~\cite{al2021financial}. In real-world scenarios, where the dataset is vast, addressing the issue of data imbalance becomes exceptionally difficult. The scale of the dataset often leads to extreme imbalances between classes, making it challenging to tackle effectively.
%Consequently, this challenge poses a potential drawback that can significantly hamper classification performance.
The dominance of the majority class tends to overshadow the classifiers' capacity to accurately classify instances from the minority class. As a result, the performance of the classifiers suffers, leading to subpar outcomes specifically for the minority class.

% Imbalanced classification: up/down sampling, but downsampling >>> upsampling, bcuz information loss. If imbalance ratio is large, more information will be loss. Existing work (AutoSMOTE) shows that
To tackle the problem, over-sampling techniques are developed to directly modify the dataset by injecting synthetic instances. One of the most representative over-sampling techniques in the literature is SMOTE~\cite{chawla2002smote}, which generates synthetic samples by performing linear interpolation between minority instances and their neighbors. 
Alternative solutions with under-sampling techniques~\cite{yen2006under,liu2020mesa} remove existing majorities from the dataset to balance the majorities and minorities. The under-sampling techniques may significantly lose information and lead to poor performances when data is extremely imbalanced. Recent study~\cite{zha2022towards} suggests that over-sampling techniques are generally favorable to extreme imbalance settings and show the promise of over-sampling techniques.

\begin{figure}
    \centering
    \includegraphics[width=1.0\linewidth]{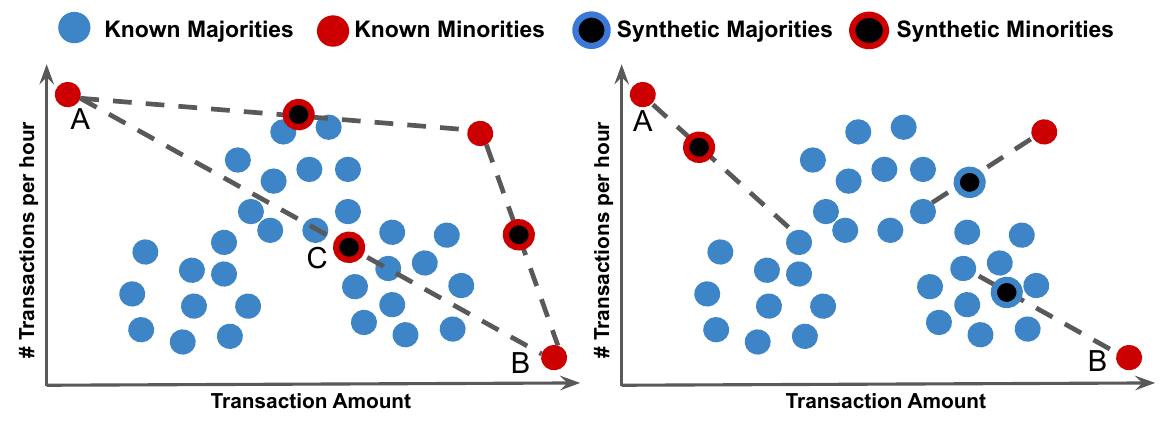}
    \caption{Comparison between synthetic oversampling (left) and mix-up (right).}
    \vspace{-25pt}
    \label{fig:illu}
\end{figure}

% In practical using sscenario, minority could be very diversed
However, in real-world application scenarios such as anomaly detection, minorities are seldom clustered within a certain distribution and may be scattered diversely in the feature space. This will lead to unsatisfactory performances due to the diverse distribution of minorities. Take the SMOTE~\cite{chawla2002smote,siriseriwan2017adaptive,nguyen2011borderline} as an example. As shown in the left side of Figure~\ref{fig:illu}, if we perform synthetic oversampling based on the abnormal accounts A and B, it is likely to create a synthetic account C labeled as abnormal but with normal transaction numbers and transaction amount. As a result, account C interleaves with the majority of normal accounts, which may degrade the classification performance.

In parallel, instead of creating synthetic samples with minorities, data mix-up~\cite{zhang2018mixup,zha2023data} techniques in the image domain~\cite{dabouei2021supermix} have achieved great success in augmenting the training data by mixing up data samples from two different classes such that fine-grained supervisory signals of label differences between the two classes can be captured. Recently, the mix-up technique has been extended to the imbalance classification~\cite{kabra2020mixboost}, which outperforms synthetic minority oversampling methods when minorities are generally within clusters. In light of the previous success, we consider synthetic instance generation based on \emph{both minority and majority instances} to address the diverse distribution of minorities. The right of Figure~\ref{fig:illu} shows how mix-ups can help. A pair of normal and abnormal accounts with a properly chosen mix-up strategy can generate not only synthetic minorities but also synthetic majorities to better shape the boundary between abnormal and normal accounts. Motivated by this, this work aims to answer the following research question: \emph{\textbf{How can we leverage the mix-up technique to better capture the limited supervisory signal from diverse minorities?}}

% \begin{figure}[t]
%   \centering
%   \begin{subfigure}[b]{0.238\textwidth}
%     \centering
%     \includegraphics[width=\textwidth]{figs/random_homo.pdf}
%     \vspace{-8pt}
%   \end{subfigure}%
%   \begin{subfigure}[b]{0.24\textwidth}
%     \centering
%     \includegraphics[width=\textwidth]{figs/random_hetero.pdf}
%     \vspace{-8pt}
%   \end{subfigure}%
%   \vspace{-8pt}
%   \caption{Comparison between random mix up anomalies (left) and random mix up anomalies with normal samples (right). Data points with black colors are synthetic samples and the edge colors are their assigned label (red as anomaly and blue as normal sample).}
%   \label{fig:prelim}
%   \vspace{-5pt}
% \end{figure}
% CH1: Select a pair of anomaly and normality
Nevertheless, it is challenging to develop a mix-up framework for the following reasons. \textbf{First}, it is challenging to identify the source samples for the mix-up.
Randomly choosing two data samples for mix-up is likely to create noise due to the extreme imbalance between minorities and majorities. The noisy samples can harm the classification performance. Additionally, even if we apply some strategies (e.g.,  only mixing up between anomalies), it is still likely to create noise due to the diverse distribution of minorities.
% CH2: Mix-up ratio
\textbf{Second}, even if we are able to get a proper pair of source samples, how to mix up remains to be a non-trivial problem. Randomly selecting a mix-up ratio may end up creating noises. We argue that the mix-up ratio needs to be tailored for the source samples. For example, if we select a pair of source samples with a majority sample in the center of majorities and a minority sample that is surrounded by majorities, a mix-up ratio that weighs more on minority samples may create a synthetic sample that better shapes the decision boundary. Conversely, for a pair of source samples with a majority sample that is deviated from most majorities and a minority sample that is nearby other minorities, a mix-up ratio that weighs more on the majority sample may lead to a more informative synthetic sample.
% CH3: Use synthetic data point as the base to select top-KNN data point as the source data point of the next iteration for exploration and use uncertainty score to encourage exploration of uncertain decision boundaries.
\textbf{Third}, even if we have a strategy to mix up data samples, simultaneously considering the underlying classifier and traversing the feature space to search for the most informative spot for synthetic sample generation is still challenging. For example, how can we know if the created sample can further benefit the formation of a model decision boundary? Likewise, if we are able to find a sweet spot for model training, how to guide the synthetic data generation process remains a challenging problem.

% as anomalies are extremely sparse, it is non-trivial to explore the potential supervisory information based on the existing labeled anomalies. For example, if we 

To tackle the challenges above, we propose MixAnN, a universal data mixer that is capable of incorporating different classifiers to explore and exploit potentially beneficial information from extremely scarse and diverse distributed minorities for training a classifier. Motivated by the success of deep reinforcement learning and its applications on imbalanced classification problem~\cite{zha2022towards,liu2020mesa}, we define an iterative mix-up process that aims at iteratively creating synthetic samples by traversing in the feature space through the nearest neighbor exploration. Then, we formulate the iterative mix-up into a Markov decision process (MDP) and design an action space to perform an automated selection of mix-up ratio with a tailored reward function that provides model uncertainty and performance improvement to guide feature space traversal. Finally, to solve the MDP, we tailor a deep actor-critic framework to tackle the discrete-continuous action space and train a mix-up policy to optimize the reward. Our contributions are summarized as follows:
\vspace{-12pt}
\begin{itemize}[leftmargin=10pt]
    \item We propose an iterative mix-up process to generalize label information by traversing the feature space.
    \item We formulate the iterative mix-up into a Markov decision process and design a combinatorial reward signal, which considers the convergence of the classifier while providing model uncertainty, to guide the mix-up process.
    \item We tailor a deep reinforcement learning algorithm to learn a mix-up policy for the discrete-continuous action space.
    \item We conduct extensive experiments to show that MixAnN outperforms the existing data augmentation approaches and state-of-the-art label-informed anomaly detectors on 7 publicly available benchmark datasets.
\end{itemize}

\section{Related Work}
% Move to last
% 1 page
Here  we briefly review the related work on anomaly detection, imbalanced classification and mix-up strategies. 

% \vspace{-5pt}
\subsection{Label-informed Anomaly Detection}
% 0.35 page
Weakly/semi-supervised anomaly detection~\cite{ienco2016semisupervised,pang2019deep,pang2019weak,zhao2018xgbod,ruff2020deep,lai2023context} are two main strategies to tackle the problem when either normal or anomaly samples are labeled. To leverage the large number of labeled normal samples, SAnDCat~\cite{ienco2016semisupervised} selects top-K representative samples from the dataset as a reference to evaluate anomaly scores based on a model learned from pair-wise distances between labeled normal instances. On the other hand, to exploit a limited number of labeled anomalies, DevNet~\cite{pang2019deep} enforces the anomaly scores of individual data instances to fit a one-sided Gaussian distribution for leveraging prior knowledge of labeled normal samples and anomalies. PRO~\cite{pang2019weak} introduces a two-stream ordinal-regression network to learn the pairwise relations between two data samples, which is assumption-free on the probability distribution of the anomaly scores.  
% However, the assumption of the Gaussian prior may suffer from diverse prior knowledge when label information is not consistently distributed. To tackle the problem, PRO~\cite{pang2019weak} introduces a two-stream ordinal-regression network to learn the pair-wise relations between two data samples, which is assumption-free on probability distribution of the anomaly scores. 

Recently, several endeavors~\cite{zhao2018xgbod,ruff2020deep} further generalize the label-informed anomaly detection problem into semi-supervised classification setting~\cite{chapelle2009semi} that limited numbers of both normal and anomaly samples are accessible. The main underlying assumptions are that similar points are likely to be of the same class and, therefore, densely clustered in low-dimensional feature space. XGBOD~\cite{zhao2018xgbod} extracts feature representation based on the anomaly score of multiple unsupervised anomaly detectors for training a supervised gradient boosting tree classifier. DeepSAD~\cite{ruff2020deep} points out that the semi-supervision assumptions only work for normal samples and further develops a one-class classification framework to cluster labeled normal samples while maximizing the distance between the labeled anomalies and the cluster in the high-dimensional space. However, weakly/semi-supervised learning methods focus on modeling the given label information without considering the relations between two labeled instances. Therefore, it is infeasible to generalize the label information when anomaly behaviors are diverse. By considering correlations between labeled samples and generating beneficial training data correspondingly, we are able to generalize the label information for training arbitrary classifiers.
\vspace{-5pt}

\subsection{Synthetic Oversampling for Imbalanced Classification}
% Synthetic oversampling cannot address anomaly distribution
% 0.35 page
Data augmentation has been extensively studied for a wide range of data types~\cite{wen2020time,shorten2019survey,ling2023learning,chen2022adversarial,song2023} to enlarge training data size and generalize model decision boundaries for improving performance and robustness. Approaches to solving the imbalanced classification problem can be categorized into two types: algorithm-wise and data-wise. Algorithm-wise approaches directly tailor the loss function of classification models~\cite{liu2006influence,cao2019learning,li2021autobalance} to better fit the data distribution. However, modifying the loss function only facilitates the fitting of label information and may suffer from generalizing label information when the behavior between minority classes varies dramatically. Data-wise approaches generate new samples into the minority class~\cite{chawla2002smote,he2008adasyn} or remove existing samples from majority classes. Synthetic Minority Oversampling (SMOTE)~\cite{chawla2002smote} generates new minority samples by linearly combining a minority sample with its k-nearest minority instances with a manually selected neighborhood size and the number of synthetic instances. A series of advancements on SMOTE~\cite{kovacs2019smote} further introduce density estimation~\cite{he2008adasyn,bunkhumpornpat2012dbsmote}, data distribution-aware sampling~\cite{han2005borderline}, and automated machine learning~\cite{zha2022towards} to tackle the class imbalance problem without manual selection of neighborhood size and the number of synthetic instances.
% \vspace{-10pt}
\vspace{-5pt}
\subsection{Mix-up}
% existing mix-up cannot be directly apply to anomaly detection
% Mixup 0.25 page
Recently, instead of conducting synthetic data sampling on a single class, Mixup~\cite{zhang2018mixup,han2022g,chen2022graph,ling2023graph} achieves a significant improvement in the image domain by synthesizing data points through linearly combining two random samples from different classes with a given combination ratio and creating soft labels for training the neural networks. As Mixup assumes that all the classes are uniformly distributed for image classification, it does not apply when the class distribution is skewed. To tackle this limitation, MixBoost~\cite{kabra2020mixboost} introduces a skewed probability distribution to sample the ratio for linearly combining two heterogeneous samples. However, the imbalanced classification problem assumes that minority samples are clustered within the feature space, which may not be true when the minorities are anomalies. To this end, our work considers the attributes of a pair of majority and minority samples for jointly identifying the best k-nearest neighborhood and combination ratios. Then, we generate the synthetic samples with the combined ratios and identify the next pair of samples within the k-nearest neighborhood. In this way, our framework is capable of exploiting the label information while exploring the diversely distributed minorities.

\begin{figure}
    \centering
    \includegraphics[width=1.0\linewidth]{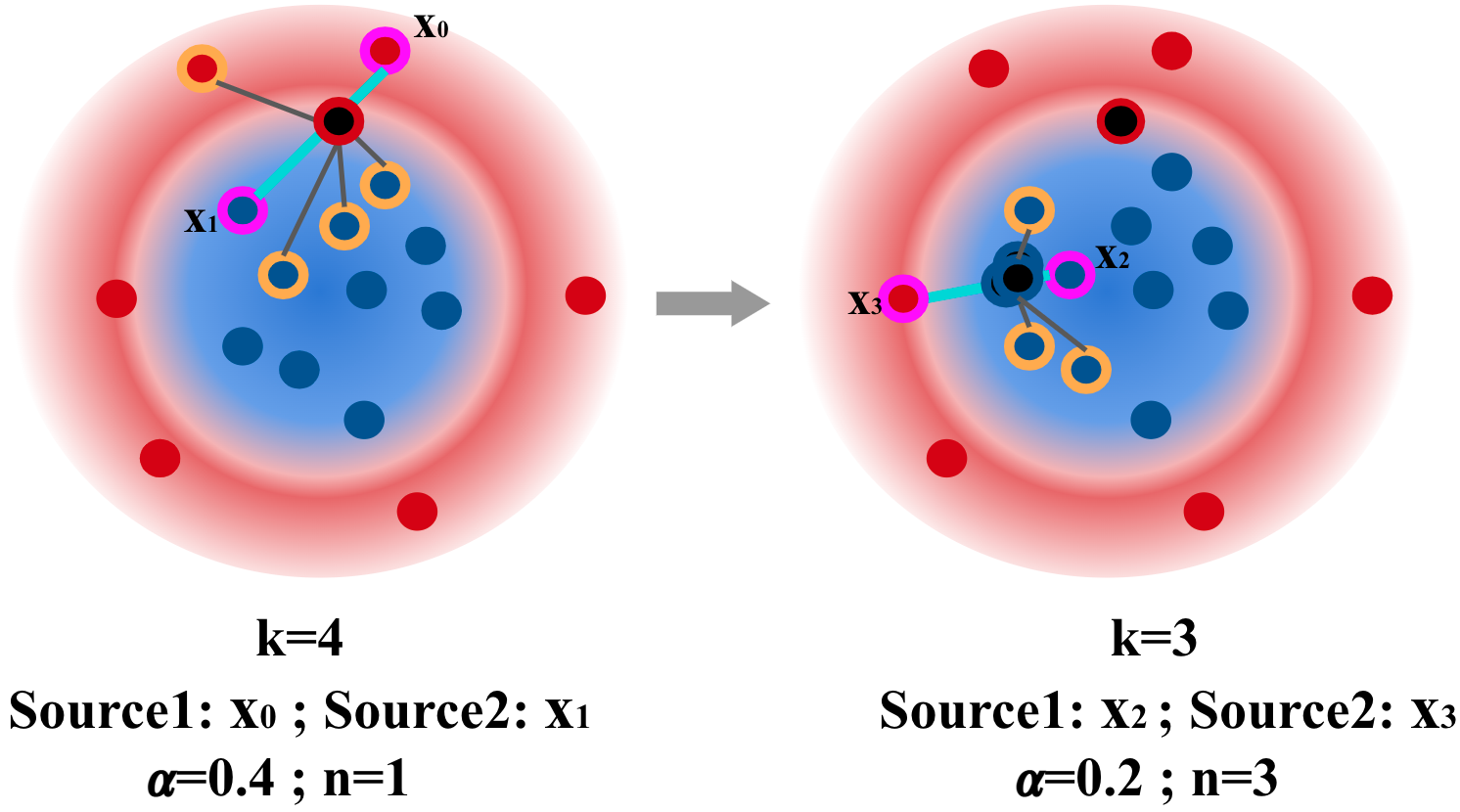}
    \vspace{-10pt}
    \caption{An illustration of the iterative mix-up process. The background colors indicate the model decision boundary. The attributes of the source samples (purple circled) are considered to specify the composition ratio $\alpha$ for oversampling the synthetic sample (black points) with corresponding labels (outer circle of black points) $n$ times. In the meantime, a $k$-nearest neighborhood is identified for randomly sampling the next pair of source samples. Finally, the process iterates to the next round with the new pair of source samples.}
    \label{fig:mdp}
    \vspace{-10pt}
\end{figure}
\section{Methodology}
\begin{figure*}
    \centering
    \includegraphics[width=0.75\textwidth]{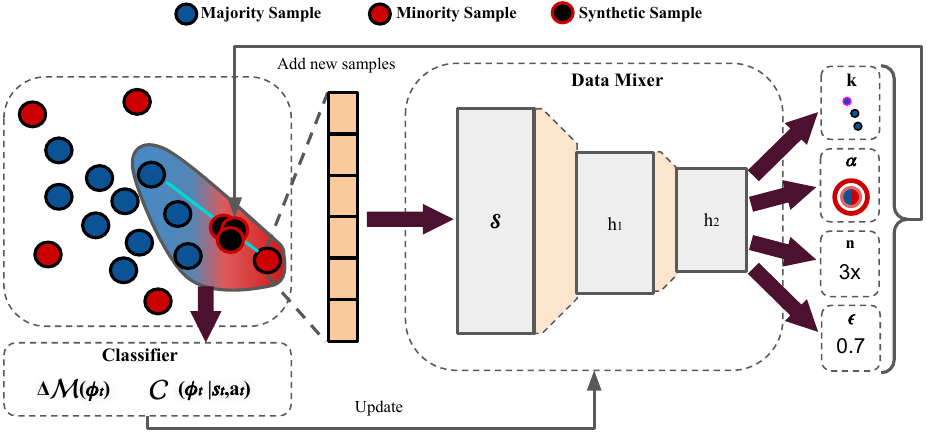}
    \vspace{-4pt}
    \caption{An illustration of our framework. In each step, a pair of a normal sample and an anomaly is inputted to the data mixer. The corresponding action (i.e., neighborhood size $k$, composition ratio $\alpha$, number of oversampling $n$, and termination probability $\epsilon$) is then generated to create synthetic samples. After that, the synthetic samples are adopted to train the classifier and yield the reward signals (performance improvement $\Delta\mathcal{M}(\phi_{t})$ and model confidence $\mathcal{C}(\phi_{t}|.)$) for updating the data mixer.}
    % {\color{red}{For the new 3 synthetic samples, can you make them more differentiable from the original pair? We could also clarify these in the annotation part. For instance, the blue one ..... is the normal sample; the right one .... is the anomaly and the three .... are new generated samples....}}
    \vspace{-4pt}
    \label{fig:framework}
\end{figure*}

% 3 pages
Figure~\ref{fig:framework} depicts an overview of our framework. We establish our framework by forming a classifier-driven Markov decision process with a data mixer that learns to generate synthetic samples for training the underlying classifier. In this section, we formally define the problem of strategic data augmentation in imbalanced classification with diversely distributed minorities. Then, we elaborate on the details of the Markov decision process, including the reward signal for iterative data mix-up and a tailored sequential decision problem solver. Finally, we provide algorithm and implementation details.

\subsection{Problem Statement}
% 0.25 page
%Following the setting of semi-supervised anomaly detection~\cite{pang2019deep,zhao2018xgbod,ruff2020deep}, 
A training dataset $\mathcal{X}^{\text{train}}=\{x_1, x_2,\dots,x_{n+m}\}$ is composed of as set of labeled majorities $\mathcal{N}=\{x_1,x_2, \dots, x_{n}\}$ and a set of labeled minorities $\mathcal{A}=x_{n+1},x_{n+2},\dots,x_{n+m}\}$, where $n \gg m$, and the minorities can be anomalies that are diversely distributed. The goal of imbalanced classification is to learn a classifier $\phi: \mathcal{X} \rightarrow \mathbb{R}$ which evaluates the probability of individual data points as majority or minority  in the given dataset $\mathcal{X}$ via exploiting the prior knowledge of labeled majorities $\mathcal{N}$ and minorities $\mathcal{A}$, so that $\sum_{i \in \mathcal{A}}{\phi(x_i)}$ 
is maximized while $\sum_{j \in \mathcal{N}}{\phi(x_j)}$ is minimized with the constraints of $\phi(x_i) > 0.5$ and $\phi(x_j) \leq 0.5$.

% so that $\phi(x_i) > 0.5$ and $\phi(x_j) \leq 0.5$ when $x_i \in \mathcal{A}$ and $x_j \in \mathcal{N}$.

To better generalize the knowledge from the label information, we formally define the problem of strategic data augmentation as follows. Given a dataset $\mathcal{X}^{\text{train}} = \{\mathcal{N},\mathcal{A}\}$ where $\mathcal{X}^{\text{train}} \in \mathbb{R}^{(n+m)\times d}$, with a supervised classifier $\phi$, we target on augmenting the $\mathcal{X}^{\text{train}}$ with a synthetic dataset $\mathcal{X}^{\text{syn}}$ according to the $\phi$, where the synthetic dataset $\mathcal{X}_{\text{syn}} \in \mathbb{R}^{l\times d}$ is generated via mixing up samples from $\mathcal{N}$ with samples from $\mathcal{A}$. Specifically, our objective is to properly sample pairs of data instances from $\mathcal{N}$ and $\mathcal{A}$ with the corresponding mix-up ratio $\alpha$ to create synthetic instances $x^{\text{syn}} \in \mathcal{X}^{\text{syn}}$, such that the performance of $\phi$ can be maximized by being trained on $\hat{\mathcal{X}}^{\text{train}} = \{\mathcal{X}^{\text{train}} \cup \mathcal{X}^{\text{syn}}\}$.

\subsection{Iterative Mix-up Process}
% 0.5page
% motivate the reason of using RL: 
% 1) Mapping from instance features to mixup criteria.
% 2) Customized feature space exploration
% 3) Iterative exploration
To generalize label information from two different classes, Mixup~\cite{zhang2018mixup} performs synthetic data generation over two samples from different classes, which has been extensively studied to augment image and textual data. The core idea of Mixup is to linearly combine two samples as follows:
\begin{equation}
    x_{\text{syn}} = \alpha * x_{0} + (1-\alpha) * x_{1},
    \label{eq:mixup_x}
\end{equation}
where $x_{0}$ and $x_{1}$ are the two selected source samples and $\alpha \in [0.0, 1.0]$  controls the composition of $x_{\text{syn}}$. Although existing works~\cite{zhang2018mixup,kabra2020mixboost} generate a soft label of $x_{\text{syn}}$ in the same fashion for the imbalance classification problem, the diverse behaviors of the anomalies lead to similar labels with high granularity on diverse synthetic samples, which may prompt the model to over-fit on noisy synthetic labels. To this end, instead of generating soft labels, we synthesize hard labels for $x_{\text{syn}}$ with a threshold $\eta$ as follows:
\begin{equation}
    y_{\text{syn}} = 
    \begin{cases}
    y_{0}, &\alpha \geq \eta, \\
    y_{1}, &\text{otherwise},
    \end{cases}
    \label{eq:mixup_y}
    \vspace{-5pt}
\end{equation}
where $0.5 \leq \eta \leq 1.0$. Due to the diverse behavior of anomalies, arbitrarily mixing up two random source samples from the $X^{train}$ may lead to noisy samples. To tackle this problem, we seek to identify a meaningful pair of samples for synthesizing new samples. As normal samples are often concentrated in the latent space~\cite{ruff2020deep}, inspired by the BorderlineSMOTE~\cite{han2005borderline} that shows that borderline samples are more informative, we propose to traverse the feature space of $X^{\text{train}}$ with the guidance of the decision boundary of the model $\phi$ for synthetic oversampling. Specifically, given a pair of arbitrary source samples, we consider the attributes of the two samples to identify corresponding $\alpha$ and number of oversampling for generating a set of $x_{syn} \in \mathcal{X}^{\text{syn}}$. Meanwhile, an optimal range for uniformly sampling the next pair of source samples is identified according to the model impact for iterating to the next round of the mix-up process. The intuition behind uniform sampling is to consider the relationship between the attributes of the source samples and their entire neighborhood information instead of focusing on a certain sample in the neighborhood. Figure~\ref{fig:mdp} illustrates an example of the iterative mix-up process. Based on the attributes of source samples $x_0$ and $x_1$, we first specify the composition ratio $\alpha=4$ and the number of oversampling $n=1$ for creating synthetic sample $x_{\text{syn}}$. Then, a $k$-nearest neighborhood is identified for sampling the next pair of source samples $x_2$ and $x_3$ for the next round of mix-up.

%  Our goal is to iteratively select $x_0$ and $x_1$ with corresponding $\alpha$ and number of samples for generating a set of $x_{syn} \in \mathcal{X}^{\text{syn}}$ during the feature space traversal while approaching the long-term goal that maximizing the performance of $\phi$ with $\mathcal{X}^{\text{syn}}$. 

The proposed iterative mix-up process has several desirable properties. First, it can make personalized decisions. For example, we can generate more samples for some instances and fewer for other instances. Second, it can incorporate various information to guide the mix-up process. For instance, data attributes and model impact can be considered and serve as guidance for generating samples. Third, by simultaneously considering the model impact with the feature distribution, we directly generate information that is missing in the original dataset but beneficial for model training.
% Third, by traversing the feature space with model status, we directly extract information that is beneficial to the model and therefore lead to meaningful synthetic data.
\vspace{-5pt}
\subsection{Formulating Iterative Mix-up as Markov Decision Process} 
The iterative mix-up process can be formulated as a Markov decision process with a quintuple $(\mathcal{S}, \mathcal{A}, \mathcal{T}, \mathcal{R}, \gamma)$, where $\mathcal{S}$ is a finite set of states, $\mathcal{A}$ is a finite set of actions, $\mathcal{T}: \mathcal{S} \times \mathcal{A} \to \mathcal{S}$ is the state transition function that maps the current state $s$, action $a$ to the next state $s'$, $\mathcal{R}: \mathcal{S} \times \mathcal{A} \to \mathbb{R}$ is the immediate reward function that reflects the quality of action $a$ for the state $s$, and $\gamma$ is a decade factor to gradually consider the future transitions. Figure~\ref{fig:framework} illustrates the Markov decision process of the framework. To sample-wise tailor the mix-up strategy, we target learning a data mixer that maps the attributes of source data samples into an augmentation strategy for creating synthetic samples while exploring model decision boundaries. The MDP can be defined as follows:

\begin{itemize}[leftmargin=10pt]
    \item \textbf{State Space ($\mathcal{S}$):} At each timestamp $t$, state $s_{t} \in \mathcal{S}$ is defined as $s_t = (x_0^{t}, x_1^{t})$, where $s_t \in \mathbb{R}^{2m}$ is a concatenation of two $m$-dimensional feature vectors of the two source samples. Therefore, the state space is defined as $\mathcal{S}=\{(x_0^{t}, x_1^{t}) | x_{0}^{t}, x_1^{t} \in \mathcal{X}^{train}\}$.
    
    \item \textbf{Action Space ($\mathcal{A}$):} At each timestamp $t$, the action $a_{t} \in \mathcal{A}$ where $a_t = (k,\alpha,n,\epsilon)$ is a vector composed of the size of neighborhood $k$, composition ratio $\alpha$, number of oversampling $n$ and the termination probability of the iterative mix-up process $\epsilon$. Therefore, the action space is defined as a discrete-continuous space $\mathcal{A}=\{ (k_t,\alpha_t,n_t,\epsilon_t) | k_t, n_t, \epsilon_t \in \mathbb{N}, \alpha_t \in \mathbb{R}^{+}\}$.

    \item \textbf{Transition Function ($\mathcal{T}$):} Given a state $s_t=(x_0^{t},x_1^{t})$ and an action $a_{t}=(k, \alpha, n, \epsilon)$, the transition function firstly adopts $\alpha$ with Eq.~\ref{eq:mixup_x} and Eq.~\ref{eq:mixup_y} to oversample $x_{\text{syn}}$ for $n$ times. Then, the resulting synthetic samples $\mathcal{X}_{\text{syn}}$ will be adopted for training the classifier and lead to a classifier $\phi_t$ in timestamp $t$. Finally, the transition function shifts to the next state $s_{t+1}=(x_0^{t+1}, x_1^{t+1})$ where the $x_0^{t+1}$ is randomly sampled from the $k$-nearest neighborhood of the $x_{\text{syn}}$ and the $x_1^{t+1}$ is identified as the nearest data point with a different label from $x_0^{t+1}$.
    \item \textbf{Reward Function ($\mathcal{R}$):} The reward signal $r_t$ for each timestamp $t$ is designed to encourage performance improvement while exploring the decision boundaries of the classifier $\phi$. Therefore, the reward function is defined as:
    \begin{displaymath}
    \mathcal{R}(s_t,a_t)=\lambda*\Delta\mathcal{M}(\phi_t)*\mathcal{C}(\phi_t|s_t,a_t),
    \end{displaymath}
    where $\lambda$ is a hyperparameter to define the strength of the reward signal, $\mathcal{M}$ is an evaluation metric, and $\Delta\mathcal{M}(\phi_t)$ measures the performance improvement of $\phi_t$. The $\mathcal{C}(\phi_t|s_t,a_t)$ evaluates the model confidence to encourage exploring the decision space of $\phi_t$. This way, the reward signal drives the data mixer to explore the classifier while achieving maximum improvement with the newly synthesized data samples. The implementation details are provided in section~\ref{sec:detail}.
    
    % P(y|s_t \in \mathcal{A}, \phi_t) P(y|s_t \in \mathcal{N}, \phi_t))
    % Specifically, the reward signal $r_t$ is defined as follows:
    %     \begin{displaymath}
    %         r_t = \mathcal{M}(\phi(\mathcal{X}^{\text{val}}), \mathcal{Y}^{\text{val}}) * (P(y|s_t, \phi) P(y|s_t, \phi))
    %     \end{displaymath}
\end{itemize}
 
% There are mainly two stages for iterating a transition with the transition function $\mathcal{T}$: target sample selection and synthetic oversampling. In the target sample selection stage, given a source sample $x_0$, the action $k$ form a candidate pool of $k$-nearest neighbors with different label from the source sample, then the target data instance $x_1$ are uniformly sampled from the candidate pool. The intuition behind the uniform sampling is to prompt the agent to consider the relationship between the attribute of the source sample and its entire neighborhood information instead of focus on a certain sample in the neighborhood. 
% In the synthetic oversampling stage, the action $\alpha$ and $n$ are incorporated with Eq.~\ref{eq:mixup_x} and Eq.~\ref{eq:mixup_y} to form a $\mathcal{X}_{\text{syn}}$ by oversampling $x_{\text{syn}}$ for $n$ times for training the model. Finally, the sampled target instance $x_0$ become the  

\subsection{Solving Markov Decision Process with Deterministic Actor-Critic}
% Motivation of using Actor-critic. 
% Transformation for Hybrid action space (continuous + discrete)
% Details of DDPG
To solve the MDP above, we define a parameterized policy $\pi_{\theta}$ as the data mixer to maximize the reward signal of the MDP where the ultimate goal is to learn an optimal policy $\pi_{\theta}^{*}$ that maximize the cumulative reward $\mathbb{E}_\pi[\sum_{t=0}^{\infty}\gamma^t r_t]$. However, the action space of the iterative mix-up process is a discrete-continuous vector, and the reward signals generated from an under-fitted $\phi_t$ may be in-stable. To this end, we employ the deep deterministic policy gradient (DDPG)~\cite{lillicrap2015continuous}, an actor-critic framework that equips with two separate networks: actor and critic. The critic network $Q_(s_t, a_t|\theta_2)$ approximates the reward signal for a state-action pair from the MDP, while the actor network $\pi_(s_t|\theta_1)$ aims to learn the policy for a given state $s_t$ based on the critic network. We note that an advanced actor-critic framework such as soft actor-critic~\cite{haarnoja2018soft} could be adopted to learn the $\pi_{\theta}^{*}$, which will be our future exploration.

% actor network
To perform a continuous action, DDPG learns an actor network $\pi(s_{t}|\theta_1)$ that deterministically maps a given state $s_t$ to an action vector $a_t$ and trains the network by maximizing the approximated cumulative reward that generated by the critic network $Q(.|\theta_2)$. Specifically, given $N$ transitions, a projected action $\pi(s_{t}|\theta_1)$ can be generated as the input of the critic to minimize the following loss function:
\begin{equation}
    \label{eq:actor}
    L_\pi(\theta_1) = -\frac{1}{N}\sum_{i=1}^{N}Q(s_i,\pi(s_{i})|\theta_2),
\end{equation}
where the action $\pi(s_{i})$ is a $4$-dimensional real number vector. To fully leverage the expressive power of the deep neural network during the training while outputting a discrete-continuous vector for the MDP, we transform the continuous action vector $\pi(s_{i})$ with a sigmoid function and yield the action vector $a_t = w\circ\sigma(\pi(s_{i}|\theta_1))$ where $w$ specifies the value constraints of individual entries. For instance, if the maximum for the $k$ and $n$ are $10$ and $5$, then $w=[10, 1, 5, 1]$ since $\alpha$ and $\epsilon$ are expected to be ranging from $0$ to $1$. The outcome for the $k$ and $n$ are rounded to the nearest integer.

% critic network
To tackle the in-stable reward signal issue, DDPG approximates the reward signal with the critic network $Q_(.|\theta_2)$ and trains the networks in an off-policy fashion. It introduces a replay buffer to store historical and randomly sampled transitions to minimize the temporal correlation between two transitions for learning across a set of uncorrelated transitions. Specifically, the critic network $Q_(.|\theta_2)$ maps a state-action pair into a real value $y_t$ via minimizing the following loss function:
\begin{equation}
    \label{eq:critic}
    L_Q(\theta_2) = [Q(s_t,a_t) - b_t]^{2}
\end{equation}
Here, $b_{t} = \mathcal{R}(s_t, a_t) + \gamma Q(s_{t+1},\pi(s_{t+1}|\theta_1)|\theta_2)$ is a signal derived from the Bellman equation~\cite{sutton2018reinforcement} which considers the recursive relation between the current reward and the future approximated reward signals for maximizing cumulative reward. $\pi(s_{t+1}|\theta_1)$ is an action from the actor network and the $\gamma$ is the decade factor.
% \begin{equation}
%     y_t = \mathcal{R}(s_t, a_t) + \gamma Q(s_{t+1},\pi(s_{t+1}|\theta_2)|\theta_1)
% \end{equation}
% where $\pi(s_{t+1}|\theta_2)$ is an action specified by the actor network and the $\gamma$ is the decade factor.

\begin{algorithm}[t]
\small
\caption{Training Procedure}
\label{xf}
\setlength{\intextsep}{0pt} 
\begin{algorithmic}[1]

\STATE \textbf{Input:} Input data $X^{\text{train}}$, maximum neighborhood size $K$, number of training episode $E$, DDPG training step $S$, reward coefficient $\lambda$, window size for baseline reward $m$.

\STATE Initialize classifier $\phi$, actor network $\pi$, critic network $Q$, memory buffer $\mathcal{B}$, $\epsilon=0$.
\STATE Split partial $X^{\text{train}}$ to form $X^{\text{val}}$ and train $\phi$ on $X^{\text{train}}$.
\STATE Pre-compute $K$-nearest neighborhood for $X^{train}$.
\STATE Randomly sample a pair of source samples to form the initial state $s_0=(x_0, x_1)$.
\FOR {$e$ = $0, 1, 2 ..., E$}
    \STATE $t \leftarrow 0$.
    \WHILE{$\epsilon < 0.5$}
        \STATE Get the action $a_t = \pi(s_t)$ from actor network.
        \STATE Get $\alpha$, $n$ from $a_t$ and adopt Eq.~\ref{eq:mixup_x}, Eq.~\ref{eq:mixup_y} to generate $x_{\text{syn}}$.
        \STATE Update $\phi$ with $x_{\text{syn}}$ and get $k$ from $a_t$.
        \STATE Obtain $k$-nearest neighborhood of $x_{\text{syn}}$.
        \STATE Form the reward $r_{t} = \lambda \Delta\mathcal{M}(.)\mathcal{C}(.)$ with $X^{val}$ and the $k$-nearest neighborhood of $x_{\text{syn}}$.
        \STATE Randomly choose next pair of source samples $(x_{0}^{t+1}, x_{1}^{t+1})$ from the $k$-nearest neighborhood of $x_{\text{syn}}$ 
        \STATE Form the next state $s_{t+1}=(x_{0}^{t+1}, x_{1}^{t+1})$ and get $\epsilon$ from $a_t$.
        \STATE Store the triplet $T_{t} = (s_{t}, a_{t}, s_{t+1}, r_{t})$ into $\mathcal{B}$.
        \STATE $t \leftarrow t+1$.
        % \IF{$\epsilon>0.5$} 
        % \STATE Break the while loop.
        % \ELSE
        % \STATE $t \leftarrow t+1$.
        % \ENDIF
        \FOR{step = $1$, $2$, .., $S$}
            \STATE Use data in $\mathcal{B}$ to update DDPG with Eq.~\ref{eq:actor} and Eq.~\ref{eq:critic}.
        \ENDFOR
    \ENDWHILE
\ENDFOR

\end{algorithmic}
\end{algorithm}

\subsection{Algorithm Details}
\label{sec:detail}
Algorithm~\ref{xf} illustrates the training procedure. We design two components to form the reward signal: improvement stimulation $\Delta\mathcal{M}(\phi_t)$ and model exploration $P(\phi_t|s_t,a_t)$. To learn an optimal policy for the target tasks, existing solutions~\cite{pang2021toward,wu2021rlad,zha2020meta} directly adopt the performance on a validation dataset as a reward signal. However, as the convergence of the underlying classifier is not guaranteed, directly learning a policy with the performance on a validation set may lead to noisy reward signals. As a result, rather than using the current model's performance $\phi_t$, we design an improvement stimulation to pursue the maximum model improvement on the validation set with a baseline performance:

\begin{equation}
\Delta\mathcal{M}(\phi_t) = \mathcal{M}(\phi_t(\mathcal{X}^{\text{val}}), y^{\text{val}}) - \frac{\sum_{i=t-m}^{t-1}\mathcal{M}(\phi_i(\mathcal{X}^{\text{val}}),y^{\text{val}})}{m-1},
\label{eq:baseline}
\end{equation}
where $\frac{\sum_{i=t-m}^{t-1}\mathcal{M}(\phi_i(\mathcal{X}^{\text{val}}),y^{\text{val}})}{m-1}$ is a baseline performance for the timestamp $t$, and $m$ controls the sample size for the baseline. 

In addition, as we aim at creating synthetic samples by mixing normal samples and anomalies while iteratively training the classifier $\phi$, it is critical to explore model decision boundaries to create beneficial samples and prevent generating noisy samples. Inspired by the confidence quantification~\cite{perini2020quantifying} for anomaly detectors, we develop a model exploration signal to quantify the instance-wise prediction uncertainty as follows:
\begin{equation}
     \mathcal{C}(\phi_t|s_t,a_t) = \frac{1}{k}\sum_{i=0}^{k} P(y_i=0|x_i,\phi_t)P(y_i=1|x_i,\phi_t),
     \label{eq:model_explore}
\end{equation}
where $x_i$ is the k-nearest neighbor of the synthesized sample in timestamp $t$ because k is the size of the nearest neighborhood specified by $a_t$. The intuition behind this is to encourage the data mixer to explore the uncertain area in the feature space.

\section{Experiment}
% 3 pages
Our experiments aim to answer the following research questions:  \textbf{RQ1}: How does the proposed framework compare against the existing data augmentation methods? \textbf{RQ2}: How does the framework compare against the existing label-informed anomaly detection methods? \textbf{RQ3}: How does each component contribute to the performance of the proposed framework? \textbf{RQ4}: How do the key hyperparameters affect the model performance? \textbf{RQ5}: How does the learned strategy of the proposed framework compare with the existing approaches? 
% \begin{itemize}[leftmargin=10pt]
%     \item \textbf{RQ1}: How does the proposed framework compare against the existing data augmentation methods? (Section 4.2)
%     \item \textbf{RQ2}: How does the framework compare against the existing label-informed anomaly detection methods? (Section 4.3)
%     \item \textbf{RQ3}: How does each component contribute to the performance of the proposed framework? (Section 4.4)
%     \item \textbf{RQ4}: How do the key hyperparameters affect the model performance? (Section 4.5)
%     \item \textbf{RQ5}: How does the learned strategy of the proposed framework compare with the existing approaches? (Section 4.6)
% \end{itemize}

\vspace{-5pt}
\subsection{Experiment Settings}
We conduct horizontal analysis to compare the proposed framework with data augmentation methods on three different classifiers. We also conduct a vertical analysis that compares the proposed framework with label-informed anomaly detectors. The details of the experimental settings are provided as follows:
\begin{table}[h]
    \centering
    \small
    % \vspace{-5pt}
    \setlength{\tabcolsep}{1.8pt}
    \begin{tabular}{l|cccc}
    \toprule
     
  & \# Samples & \# Features & \% Anomaly & Domain \\
 
    \midrule
    \midrule
     Japanese Vowels & 1,456 & 12 & 3.43\% & Utterance \\
     Annthyroid & 7,200 & 22 & 7.42\% & Clinical Record \\
     Mammography & 11,183 & 6 & 2.32\% & Medical Image \\
     Satellite & 6,435 & 36 & 31.63\% & Remote Sensing \\
     SMTP & 95,156 & 41 & 0.03\% & Server Log \\
     CIFAR-10 & 5,263 & 512 & 5.00\% & General \\
     FashionMNIST & 6,315 &  512 & 5.00\% & Fashion \\
     \bottomrule
    \end{tabular}
    \caption{Dataset statistics.}
    \label{tab:stats}
    \vspace{-25pt}
\end{table}

\noindent\textbf{Datasets:} Table~\ref{tab:stats} summarizes the statistics of the datasets. We adopt 5 benchmark datasets from different domains. \textbf{Japanese Vowels} contains utterances of /ae/that were recorded from nine speakers with 12 LPC cepstrum coefficients. The goal is to identify the outlier speaker.  \textbf{Annthyroid} is a set of clinical records that record 21 physical attributes of over 7200 patients. The goal is to identify the patients that are potentially suffering from hypothyroidism. \textbf{Mammography} is composed of 6 features extracted from the images, including shape, margin, density, etc. The goal is to identify malignant cases that could potentially lead to breast cancer. \textbf{Satellite} contains the remote sensing data of 6,435 regions, where each region is segmented into a 3x3 square neighborhood region and is monitored by 4 different light wavelengths captured from the satellite images of the Earth. The goal is to identify regions with abnormal soil status. \textbf{SMTP} has 95,156 server connections with 41 connection attributes, including duration, src\_byte, dst\_byte, and so on. The task is to identify malicious attacks from the connection log. \textbf{CIFAR-10 and FashionMNIST} are the benchmark datasets. We follow the setting of previous works~\cite{han2022adbench,ruff2020deep} to set one of the classes as normal and downsample the rest classes to 5\% of the total instances as anomalies. We report the average performance of airplane and automobile for CIFAR-10; and the average performance of t-shirt/top and trouser for FashionMNIST.

\noindent\textbf{Horizontal Baselines:}
We conduct a horizontal comparison between the proposed methods and the following representative data augmentation methods on 3 different classifiers (i.e., KNN, XGBoost, MLP): \textbf{Random} is a basic baseline which generates synthetic anomalies by randomly averaging two existing anomalies. \textbf{SMOTE}~\cite{chawla2002smote} linearly combines existing anomalies with their K-nearest anomalies through randomly sampled combination ratios. \textbf{BorderlineSMOTE}~\cite{han2005borderline} identifies a borderline between anomalies and normal samples with the K-nearest neighborhood of each anomaly. Then, SMOTE is performed on the anomalies in the borderline area. \textbf{SVMSMOTE}~\cite{nguyen2011borderline} introduces a support vector classifier to identify a borderline between anomalies and normal samples and perform SMOTE on anomalies near to the borderline. \textbf{ADASYN}~\cite{he2008adasyn} calculates the number of synthetic samples generated from individual anomalies using the estimated local distribution and then performs SMOTE on them. \textbf{AutoSMOTE}~\cite{zha2022towards} performs SMOTE by exploiting deep reinforcement learning to automatically select the anomalies and the corresponding oversampling strategies. \textbf{MixBoost}~\cite{kabra2020mixboost} uses a randomly sampled combination ratio to mix anomalies and normals. The number of oversamples for individual anomalies is weightily sampled based on the entropy of the underlying classifier. 
% \begin{itemize}[leftmargin=10pt]
%     \item \textbf{Random} is a basic baseline which generates synthetic anomalies by randomly averaging two existing anomalies.
%     \item \textbf{SMOTE} linearly combines existing anomalies with their K-nearest anomalies through randomly sampled combination ratios.
%     \item \textbf{BorderlineSMOTE} identify a borderline between anomalies and normal samples with the K-nearest neighborhood of each anomaly. Then, the SMOTE is performed on the anomalies in the borderline area.
%     \item \textbf{SVMSMOTE} introduces a support vector classifier to identify a borderline between anomalies and normal samples and perform SMOTE on anomalies near to the borderline.
%     \item \textbf{ADASYN} identify the number of synthetic samples generated from individual anomalies based on the estimation of the local distribution and perform SMOTE correspondingly.
%      \item \textbf{MixBoost} randomly mix up anomalies with normalities with a random sampled combination ratio. The number of oversampling for individual anomalies is weightily sampled based on entropy of the underlying classifier. This can be deemed as the proposed framework with a random mixture policy.
% \end{itemize}

\noindent\textbf{Vertical Baselines:} We perform a vertical comparison between the proposed method and label-informed detection algorithms to study the effectiveness of synthetic oversampling. \textbf{XGBOD}~\cite{zhao2018xgbod} exploits the anomalous scores of individual data points generated from multiple unsupervised algorithms as the input feature for the underlying XGBoost classifier. \textbf{DeepSAD}~\cite{ruff2020deep} develops a semi-supervised loss function that classifies the known normal samples and unlabeled samples into a unified cluster and deviates the known anomalies from the cluster. \textbf{DevNet}~\cite{pang2019deep} uses label information with a prior probability to enforce significant deviations in anomaly scores from the majority of normal samples.
% \begin{itemize}[leftmargin=10pt]
%     \item \textbf{XGBOD} exploits the anomalous scores of individual data points generated from multiple unsupervised algorithms as the input feature for the underlying XGBoost classifier.
%     \item \textbf{DeepSAD} 
%     \item DevNet
% \end{itemize}

\noindent\textbf{Evaluation Protocol:} We adopt macro-averaged precision, recall, and F1-score, which compute the scores separately for each class and average them. The intuition behind this is to equalize the importance of anomaly detection and normal sample classification since the minimum false alarms are also a critical evaluation criterion. We use 80\% data for training and 20\% for testing, where 20\% of the training data is further split into a validation set for our framework to generate reward signals or for baseline methods to perform model tuning. We run the experiments with 5 random seeds and report the average performance on the testing set.

\noindent\textbf{Implementation Details:} For the horizontal analysis, we use a KNN classifier with $k=10$, an XGBoost classifier with the linear kernel, and the Adam optimizer with ReLU activation function for a 128-64 multi-layer perceptron classifier. For the vertical analysis, we adopt XGBOD~\footnote{\url{https://github.com/yzhao062/pyod}} and public available implementations of DevNet~\footnote{\url{https://github.com/GuansongPang/deviation-network}} and DeepSAD~\footnote{\url{https://github.com/lukasruff/Deep-SAD-PyTorch}}. Since the output of Dev and DeepSAD are anomaly scores, we search for the thresholds for the two methods from $\{0.5x, 1.0x, 1.5x, 2.0x\}$ of the anomaly ratio to perform classification and report the best result. For our own method, we select the maximum neighborhood size $K$ from $\{5, 10, 15, 20, 25\}$, and set $\eta=0.3$, the reward coefficient $\lambda=10.0$, the window size for baseline $m=25$ and adopt the macro-averaged F1-score for the reward signal $\Delta\mathcal{M}$. 
% More details can be found in Appendix.

\vspace{-7.5pt}
\subsection{Horizontal Analysis}
To answer \textbf{RQ1}, we compare the MixAnN to cutting-edge synthetic oversampling methods for anomaly detection. Table~\ref{tab:horizon} tabulates the macro-averaged precision, recall, and F1-score of each data augmentation method across 3 different classifiers. We also report the performances without data augmentation on the original classifiers to show insights into how different classifiers impact the performances. In general, the proposed MixAnN is able to outperform all of the baseline oversampling methods and achieve at least $7.6\%$ and at most $33.3\%$ improvements on the F1-score. Based on Table~\ref{tab:horizon}, we make the following observations.

First, by comparing the baseline augmentation methods with the classifiers without data augmentation, we observe that the performance of the baseline augmentation methods is generally inferior to the classifier trained without data augmentation. Specifically, the average F1-scores of the baseline augmentation methods are consistently lower than the vanilla classifiers on the five datasets. The only exception is the Mixboost with KNN classifier, which is due to its decent performance on the SMTP dataset. Our further investigation into this phenomenon suggests that randomly mixing up normal samples with anomalies can create beneficial synthetic normal samples that concrete the decision boundary when anomalies are extremely sparse. This supports our claim that the existing data augmentation methods are not capable of handling the diverse behavior of anomalies and may lead to noisy synthetic samples. But generalizing label anomaly information by mixing normal samples with anomalies could alleviate the problem.

%  Observation1: V.S w/o Augmentation --> helpful for all classifiers. which classifier is more suitable for MixAnN?

Second, by comparing the MixAnN with the vanilla classifier without data augmentation, we observe that the MixAnN consistently outperforms all vanilla classifiers. On the five datasets, the MixAnN improves the F1-score of KNN, XGBoost, and MLP classifiers by $9.6\%$, $14.2\%$, and $11.5\%$, respectively. This phenomenon suggests that the MixAnN is able to adaptively create synthetic samples for different classifiers toward performance improvements. In addition, we also observe that the KNN classifier has the maximum improvement, which suggests that the nearest-neighbor exploration of our transition function favors the classifier with similar attributes. Another interesting observation is that the more complex the models, the fewer the improvements. A possible explanation is that complex models tend to be overconfident in the prediction~\cite{nguyen2015deep}, which may lead to noisy prediction uncertainty reward and mislead the learning procedure of the augmentation strategy. We will study this issue in the future.

% Observation2: V.S Mixboost --> Policy is learned
Third, by comparing the MixAnN with all other data augmentation methods, we observe that the MixAnN outperforms all baselines with the three classifiers on Macro-F1 scores. Specifically, the MixAnN averagely outperforms the F1-score of the second-best augmentation method with KNN, XGBoost, and MLP classifiers on the seven datasets by $4.9\%$, $10.1\%$ and $7.4\%$, respectively. Because Mixboost is similar to MixAnN with a random mixture policy, it implies that the proposed framework can learn tailored mix-up policies for different classifiers and data samples. We can also observe that, although the MixAnN is not always superior to all other baselines on precision and recall, the F1-scores are always the best. This phenomenon suggests that MixAnN is able to balance the trade-off between precision and recall, and therefore leads to superior F1-scores in all settings. The main reason behind this is that we adopt the Macro-F1 to form a reward signal. One may also tailor their own metrics (e.g., precision, recall, or tailored metrics) to obtain an anomaly detector that meets their requirements.

%  Observation4: V.S SVMSMOTE/BorderlineSMOTE --> uncertainty encourage reward is helpful.
Fourth, by taking a detailed comparison between the MixAnN with SVMSMOTE and BorderlineSMOTE, we observe that the F1-score of the MixAnN is superior to the two baselines by at least $14.9\%$, $14.2\%$ and $14.5\%$ with the three classifiers respectively. This phenomenon suggests that the instance-wise prediction uncertainty in the reward function is a better approach to generating tailored beneficial synthetic samples for different classifiers. The rationale behind this is that the two baselines identify the class boundary in the label space and the hyperspace of the SVM, where the MixAnN identifies the boundary that is directly defined by the underlying classifier. By encouraging the policy to generate samples on the boundary defined by the classifier, it is more likely to create beneficial information that cannot be observed from the original feature space or the hyperspace of another classifier.
\begin{table*}[h]
\centering
\small
\setlength{\tabcolsep}{2.0pt}
\begin{tabular}{l|l|ccccccc|c}
% \begin{tabular}{l|l|BBBBBBB|BB}
\toprule
\multirow{2}{*}{Classifier} & Methods & \multicolumn{7}{c|}{Dataset} & \multirow{2}{*}{Avg. F1 $\uparrow$}\\
\cline{3-9}
~ & (Prec./Rec./F1) & Vowels & Annthyroid & Mammography & Satellite  & SMTP & CIFAR-10 & FashionMNIST \\
\midrule
\midrule
\multirow{9}{*}{KNN} & w/o Augmentation & 0.98/0.72/0.82 & 0.93/0.61/0.67 & 0.86/0.74/0.79 & 0.90/0.88/0.89 & 1.00/0.75/0.83 & 0.92/0.54/0.56 & 0.96/0.82/0.87 & +9.6\% \\
~ & Random  & 0.87/0.82/0.84 & 0.67/0.63/0.65 & 0.64/0.77/0.68 & 0.90/0.88/0.89 & 0.51/0.99/0.51 & 0.55/0.55/0.55 & 0.85/0.85/0.85 & +19.7\% \\
~ & SMOTE  & 0.82/0.99/0.88 & 0.61/0.70/0.61 & 0.65/0.91/0.71 & 0.89/0.88/0.89 & 0.52/1.00/0.53 & 0.56/0.74/0.56 & 0.81/0.91/0.85  & +18.1\% \\
~ & BorderlineSMOTE & 0.85/0.99/0.91 & 0.61/0.72/0.63 & 0.67/0.92/0.73 & 0.90/0.88/0.89 & 0.52/1.00/0.53 & 0.58/0.75/0.60 & 0.81/0.91/0.86 & +14.9\% \\
~ & SVMSMOTE  & 0.85/0.99/0.91 & 0.64/0.72/0.66 & 0.70/0.92/0.76 & 0.90/0.88/0.89 & 0.53/1.00/0.56 & 0.59/0.72/0.62 & 0.84/0.91/0.87 & +13.3\% \\
~ & ADASYN  & 0.82/0.99/0.88 & 0.59/0.71/0.60 & 0.63/0.92/0.69 & 0.90/0.88/0.89 & 0.55/1.00/0.60 & 0.56/0.73/0.55 & 0.79/0.92/0.83 & +19.7\%\\ 
~ & AutoSMOTE  & 0.94/0.94/0.94 & 0.67/0.75/0.70 & 0.78/0.85/0.81 & 0.86/0.90/0.89 & 0.86/0.90/0.87 & 0.62/0.62/0.62 & 0.89/0.87/0.88 & +4.9\% \\ % AutoSMOTE
~ & Mixboost  & 0.99/0.75/0.83 & 0.94/0.65/0.72 & 0.86/0.77/0.81 & 0.90/0.88/0.89 & 0.77/1.00/0.85 & 0.56/0.56/0.56 & 0.94/0.81/0.86 & +7.6\% \\\cmidrule{2-10}
~ & MixAnN & \textbf{0.99}/0.93/\textbf{0.96} & \textbf{0.96}/\textbf{0.72}/\textbf{0.75} & \textbf{0.92}/0.80/\textbf{0.84} & \textbf{0.92}/\textbf{0.90}/\textbf{0.91} & 0.93/0.99/\textbf{0.96} & 0.84/0.58/\textbf{0.62} &\textbf{0.97}/0.83/\textbf{0.88} & - \\\midrule\midrule

\multirow{9}{*}{XGBoost} & w/o Augmentation & 0.99/0.86/0.91 & 0.94/0.67/0.73  & 0.86/0.76/0.80 & 0.90/0.76/0.79 & 1.00/0.58/0.64 & 0.69/0.66/0.67 & 0.87/0.87/0.87 & +14.2\% \\
~ & Random  & 0.92/0.89/0.90 & 0.91/0.70/0.76 & 0.66/0.75/0.70  & 0.91/0.80/0.83  & 0.50/0.49/0.49  & 0.62/0.65/0.63 & 0.64/0.72/0.68 & +23.9\%  \\
~ & SMOTE  & 0.78/0.98/0.85 & 0.86/0.90/0.88 & 0.59/0.90/0.62 & 0.91/0.80/0.84 & 0.55/1.00/0.59  & 0.62/0.67/0.64 & 0.84/0.90/0.87 & +15.8\%  \\
~ & BorderlineSMOTE & 0.79/0.98/0.86 & 0.87/0.92/0.89 & 0.58/0.90/0.60 & 0.86/0.81/0.83 & 0.50/0.94/0.47  & 0.63/0.66/0.64 & 0.87/0.88/0.87 & +18.9\%  \\
~ & SVMSMOTE  & 0.82/0.99/0.88 & 0.86/0.87/0.86 & 0.63/0.92/0.68 & 0.84/0.81/0.82 & 0.56/1.00/0.60  & 0.64/0.66/0.65 & 0.86/0.89/0.87 & +14.2\%  \\
~ & ADASYN  & 0.75/0.97/0.82  & 0.84/0.96/0.89 & 0.56/0.87/0.55 & 0.85/0.81/0.82  & 0.50/0.77/0.35  & 0.63/0.67/0.65 & 0.83/0.89/0.86 &  +23.9\%  \\ 
~ & AutoSMOTE  & 0.90/0.96/0.93 & 0.88/0.92/0.90 & 0.80/0.82/0.81 & 0.68/0.70/0.62 & 1.00/0.75/0.83 & 0.64/0.67/0.66 & 0.87/0.90/0.88 & +10.1\% \\ % AutoSMOTE
~ & Mixboost  & 0.99/0.79/0.86 & 0.96/0.55/0.57 & 0.90/0.62/0.68 & 0.89/0.70/0.73 & 0.50/0.50/0.50  & 0.67/0.61/0.63 & 0.66/0.70/0.68 &  +33.3\%  \\\cmidrule{2-10}
~ & MixAnN  & \textbf{1.00}/0.96/\textbf{0.98} & 0.87 /\textbf{0.97}/\textbf{0.92} & 0.81/0.85/\textbf{0.83} & \textbf{0.91}/\textbf{0.82}/\textbf{0.84} & \textbf{1.00}/\textbf{1.00}/\textbf{1.00}  & \textbf{0.71}/\textbf{0.68}/\textbf{0.70} & \textbf{0.89}/\textbf{0.90}/\textbf{0.90} & - \\\midrule\midrule

\multirow{9}{*}{MLP} & w/o Augmentation & 0.95/0.89/0.92 & 0.94/0.53/0.53 & 0.88/0.76/0.81 & 0.84/0.78/0.80 & 1.00/0.75/0.83  & 0.86/0.65/0.71 & 0.97/0.85/0.89  & +11.5\% \\
~ & Random & 0.78/0.88/0.82 & 0.91/0.58/0.61 & 0.62/0.77/0.66 & 0.90/0.79/0.82 & 0.50/0.98/0.50 & 0.70/0.68/0.69 &  0.83/0.87/0.85 & +22.5\%\\
~ & SMOTE  & 0.85/0.99/0.91 & 0.59/0.73/0.59 & 0.62/0.92/0.67 & 0.88/0.81/0.83 & 0.51/0.99/0.51 & 0.73/0.69/0.71 & 0.88/0.92/0.90  & +19.2\%\\
~ & BorderlineSMOTE & 0.82/0.99/0.88 & 0.59/0.73/0.59 & 0.64/0.92/0.70 & 0.85/0.80/0.82 & 0.50/0.97/0.49  & 0.73/0.71/0.72 & 0.89/0.92/0.90 & +19.2\% \\
~ & SVMSMOTE  & 0.89/0.99/0.93 & 0.62/0.74/0.65 & 0.71/0.93/0.78 & 0.88/0.81/0.83 & 0.51/0.99/0.53  & 0.75/0.70/0.72 & 0.96/0.87/0.90 & +14.5\%\\
~ & ADASYN  & 0.89/0.99/0.93 & 0.59/0.74/0.59 & 0.61/0.92/0.65 & 0.88/0.81/0.84 & 0.50/0.98/0.50  & 0.71/0.69/0.70 & 0.87/0.92/0.90 & +19.2\%\\ 
~ & AutoSMOTE  & 0.96/0.93/0.94 & 0.90/0.64/0.70 & 0.78/0.86/0.82 & 0.70/0.73/0.66 & 1.00/0.92/0.95 & 0.74/0.79/0.73 & 0.94/0.90/0.91 & +7.4\% \\ % AutoSMOTE
~ & Mixboost  & 0.65/0.60/0.62 & 0.96/0.52/0.52 & 0.90/0.72/0.78 & 0.89/0.73/0.76 & 0.90/0.83/0.86  & 0.67/0.58/0.61 & 0.93/0.95/0.88 & +20.8\% \\\cmidrule{2-10}
~ & MixAnN  & \textbf{0.99}/0.95/\textbf{0.96} & 0.93/0.74/\textbf{0.79} & 0.87/0.84/\textbf{0.85} & \textbf{0.91}/0.83/\textbf{0.85} & \textbf{1.00}/0.92/\textbf{0.95}  & 0.84/\textbf{0.71}/\textbf{0.76} & \textbf{0.97}/0.91/\textbf{0.93} & - \\
\bottomrule
\end{tabular}
\caption{Horizontal comparison with data-augmentation methods}
\label{tab:horizon}
\vspace{-10pt}
\end{table*}

\begin{table*}[h]
\centering
\small
 \setlength{\tabcolsep}{1.2pt}
\begin{tabular}{l|l|ccccccc|c}
\toprule
\multirow{2}{*}{Category} & Methods & \multicolumn{7}{c|}{Dataset} & \multirow{2}{*}{Avg. F1 $\uparrow$} \\
\cline{3-9}
~ & (Prec./Rec./F1) & Vowels & Annthyroid & Mammography & Satellite  & SMTP & CIFAR-10 & FashionMNIST & ~\\
\midrule
\midrule
% \multirow{3}{*}{Original} 
% % & LogisticRegression  & 0.96/0.93/0.94 & 0.96/0.51/0.50 & 0.86/0.76/0.80 & 0.90/0.82/0.84 & 1.00/0.75/0.83 & \\
%  & KNN  & 0.98/0.57/0.61 & 0.93/0.61/0.67 & 0.86/0.74/0.79 & 0.90/0.88/0.89 & 1.00/0.75/0.83 &  \\
% ~ & XGBoost  & 0.99/0.86/0.91 & 0.94/0.67/0.73  & 0.86/0.76/0.80 & 0.90/0.76/0.79 & 0.50/0.50/0.50 & \\
% ~ & MLP  & 0.95/0.89/0.92 & 0.94/0.53/0.53 & 0.88/0.76/0.81 & 0.84/0.78/0.80 & 1.00/0.75/0.83 & \\ \midrule
\multirow{3}{*}{Label-informed} & XGBOD & 0.86/0.71/0.76 & 0.88/0.52/0.52 & 0.95/0.59/0.65 & 0.91/0.80/0.84 & 1.00/0.58/0.64 & 0.87/0.59/0.63 & 0.96/0.83/0.88 & +28.2\%  \\ % 0.71
~ & DevNet & 0.89/0.99/0.93 & 0.63/0.61/0.62 & 0.72/0.92/0.79 & 0.71/0.71/0.71 & 1.00/1.00/1.00 & 0.72/0.74/0.73 & 0.91/0.91/0.91 & +12.3\% \\ % 0.81
~ & DeepSAD & 0.99/0.71/0.80 & 0.90/0.56/0.59 & 0.66/0.65/0.65 & 0.81/0.73/0.75 & 0.99/0.71/0.80 & 0.71/0.73/0.72 & 0.92/0.93/0.93 & +21.3\% \\\midrule % 0.75
\multirow{2}{*}{Data Augmentation} ~ & Best Baseline & 0.96/0.93/0.94 & 0.88/0.92/0.90 & 0.78/0.86/0.82 & 0.90/0.88/0.89 & 0.86/0.90/0.87 & 0.74/0.79/0.73 & 0.94/0.90/0.91  & +5.8\% \\ % 0.86
~ & MixAnN & \textbf{1.00}/0.96/\textbf{0.98} & 0.87/\textbf{0.97}/\textbf{0.92} & 0.87/0.84/\textbf{0.85} & \textbf{0.92}/\textbf{0.90}/\textbf{0.91} & \textbf{1.00}/\textbf{1.00}/\textbf{1.00} & 0.84/0.71/\textbf{0.76} & \textbf{0.97}/0.91/\textbf{0.93}  & -\\ % 0.91
\bottomrule
\end{tabular}
\caption{Vertical comparison with label-informed methods. The best baseline in the data augmentation category refers to the horizontal baseline with the best F1-score across the three classifiers.}
\label{tab:vertical}
\vspace{-10pt}
\end{table*}

\vspace{-12.5pt}
\subsection{Vertical Analysis}
In order to answer \textbf{RQ2}, we compare the MixAnN to the most advanced label-informed anomaly detection algorithms and baseline augmentation methods. Table~\ref{tab:vertical} presents the macro-averaged precision, recall, and F1-score of each method on the 5 datasets. "Best baseline" refers to the horizontal baseline with the best F1-score on individual datasets. We make the following two observations.
% Obs1: Data Augmentation in general better than label-informed.
% Label information generalization > directly using label information.

First, data augmentation methods generally outperform label-informed anomaly detectors. Comparing the best data augmentation baseline to the three label-informed approaches, the best data augmentation baseline outperforms the best label-informed algorithm by $6.2\%$. This suggests that data augmentation may be more effective in generalizing label information when incorporated with proper strategy. Additionally, the proposed MixAnN with a properly learned strategy achieves superior performance, which further validates the suggestion above.
% Obs2: High precision low recall on label-informed methods.
% Precision: Labeled-informed > best baseline, better exploits label information but over-fit on the label inforation and therefore low recall.
% MixAnN achieve comparable precision with significantly higher recall 

Second, label-informed methods achieve better precision and lower recall. By comparing  Table~\ref{tab:vertical} with Table~\ref{tab:horizon}, we can observe that the average precision of label-informed approaches is generally superior to data augmentation methods on all three classifiers, whereas the average recall is generally inferior to those baselines. One possible explanation is that label-informed approaches can only exploit the labels themselves, while data augmentation methods are capable of exploring potentially beneficial information from the limited label information. This suggests that label-informed approaches tend to over-fit existing labels, which may be sub-optimal.

\begin{table}[h]
    \centering
    \small
    \begin{tabular}{l|ccc}
    \toprule
     
    Ablations & Precision & Recall & F1-score \\
    \midrule
    \midrule
     Random reward & 0.85 & 0.60 & 0.66 \\
     w/o Improvement Stimulation (Eq.~\ref{eq:baseline})  & 0.98 & 0.64 & 0.71 \\
     w/o Model Exploration (Eq.~\ref{eq:model_explore})  & 0.98 & 0.71 & 0.79 \\
     \midrule
     Full MixAnN & \textbf{0.99} & \textbf{0.93} & \textbf{0.96} \\

     \bottomrule
    \end{tabular}
     \caption{Macro-averaged scores of MixAnN with the KNN classifier and the ablations on Japanese Vowels.}
     \label{tab:ablations}
     \vspace{-10pt}
\end{table}
\subsection{Ablation Study}
To answer \textbf{RQ3}, we conducted an ablation study on the Japanese Vowels dataset with the KNN classifier. As the reward signal is the most critical guidance toward optimal augmentation strategies, we ablate the reward function to study the contribution of individual components. Note that a random reward generates a reward signal with a random floating number from 0 to 1, which potentially leads to a random augmentation strategy. From Table~\ref{tab:ablations}, we can make the following observations.

First, the proposed MDP is solvable, and the tailored RL agent is capable of learning an optimal strategy. By comparing the random reward baseline with the MixAnN, we observe that there are significant improvements in all three metrics. Both the two ablations on Eq.~\ref{eq:baseline} and ~\ref{eq:model_explore} are significantly better than the random reward baseline, which suggests that the tailored RL agent is capable of addressing the MDP toward an optimal augmentation strategy.

Second, the two components in the proposed reward signal play significant roles in an optimal augmentation strategy. Specifically, both components are capable of increasing the exploitation of the label information and therefore lead to significant improvements in precision. On one hand, as the classifier may suffer from under-fitting during the training procedure, learning the augmentation strategy without Eq.~\ref{eq:baseline} may lead to a significant performance drop. On the other hand, without considering the model impact via Eq.~\ref{eq:model_explore}, it is less possible to identify potentially beneficial information for the underlying classifier and therefore leads to a lower recall. 

% \vspace{-5pt}
\begin{figure}[h]
  \centering
    \begin{subfigure}[b]{0.25\textwidth}
    \centering
    \includegraphics[width=0.99\textwidth]{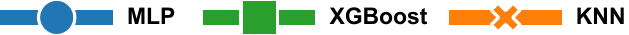}
    \vspace{-10pt}
  \end{subfigure}%
  
  \begin{subfigure}[b]{0.24\textwidth}
    \centering
    \includegraphics[width=\textwidth]{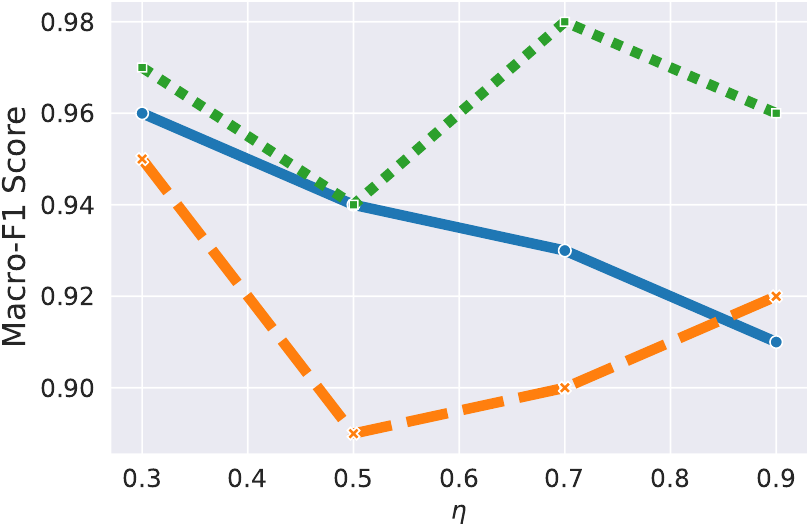}
    \vspace{-8pt}
  \end{subfigure}%
  \hspace{1pt}
  \begin{subfigure}[b]{0.23\textwidth}
    \centering
    \includegraphics[width=\textwidth]{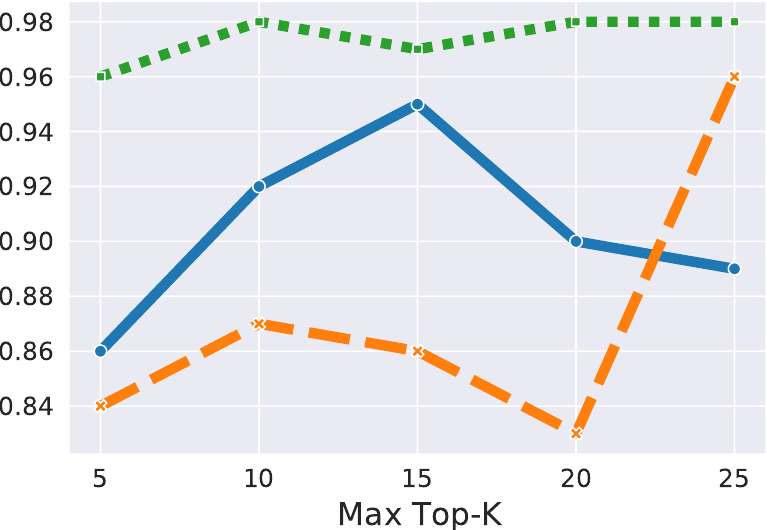}
    \vspace{-8pt}
  \end{subfigure}%
  \vspace{-10pt}
  \caption{Hyperparameter study on Vowels.}
  \label{fig:hyper}
  \vspace{-5pt}
\end{figure}
\vspace{-8pt}
\subsection{Hyper-parameter Study}
% top-k and \eta
To answer the \textbf{RQ4}, we study the two key hyperparameters (i.e., maximum neighborhood size $K$ and the threshold $\eta$) on the vowels dataset with the three classifiers. Based on the results shown in Figure~\ref{fig:hyper}, we make the following observations.

For the size of the neighborhood $K$, simpler classifiers may require collecting a larger amount of neighborhood information to extend the supervisory information of the source samples. For a complex model such as a multi-layer perceptron, we may need to carefully select the size to prevent over-smoothing when collecting too much neighborhood information~\cite{chen2020measuring} for the classifier. As for the threshold $\eta$, we observe that a lower threshold for creating synthetic anomalies generally leads to better performance. One possible explanation is that a lower $\eta$ prompts the data mixer to create more synthetic anomalies, which is therefore beneficial to the extreme imbalance setting of anomaly detection.

\begin{figure}[h]
  \centering
  \begin{subfigure}[b]{0.35\textwidth}
    \centering
    \includegraphics[width=0.99\textwidth]{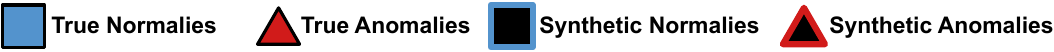}
    \vspace{-10pt}
  \end{subfigure}%

  \begin{subfigure}[b]{0.16\textwidth}
    \centering
    \includegraphics[width=0.99\textwidth]{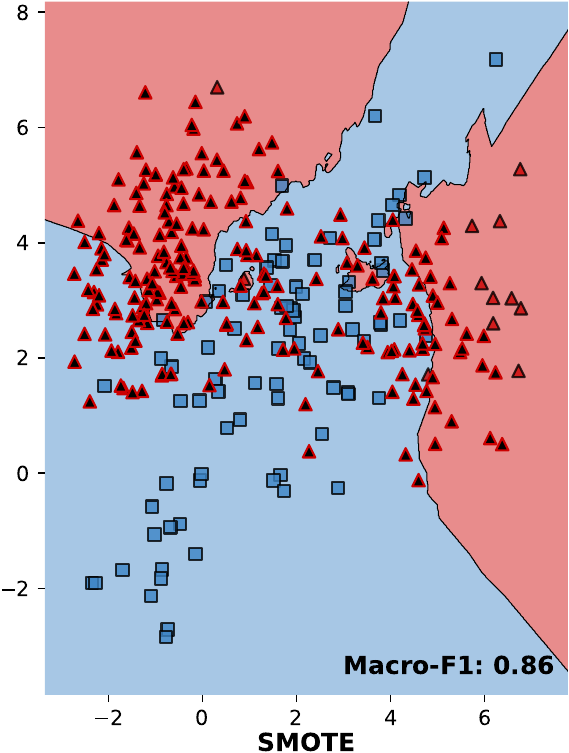}
    \vspace{-10pt}
  \end{subfigure}%
  \begin{subfigure}[b]{0.16\textwidth}
    \centering
    \includegraphics[width=0.99\textwidth]{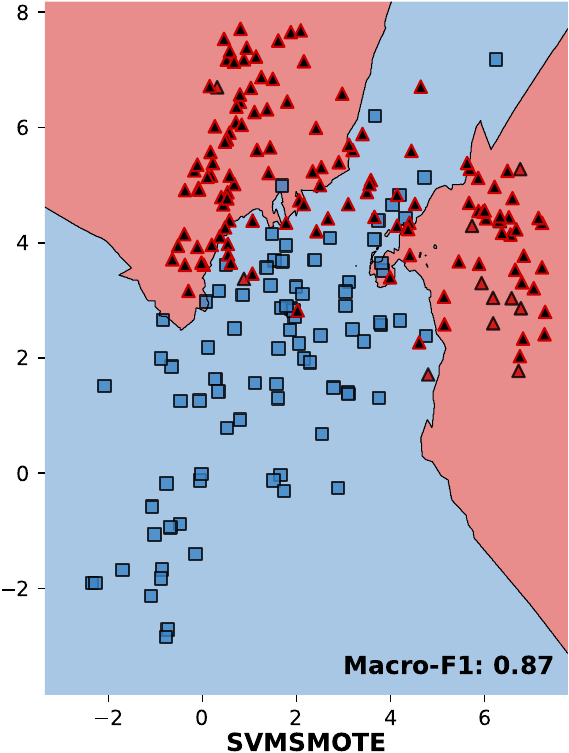} 
    \vspace{-10pt}
  \end{subfigure}%
  \begin{subfigure}[b]{0.16\textwidth}
    \centering
    \includegraphics[width=0.99\textwidth]{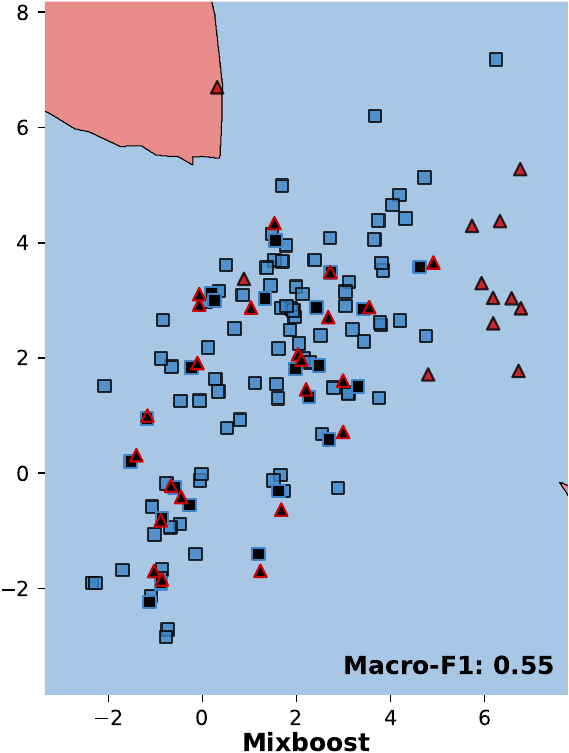}
    \vspace{-10pt}
  \end{subfigure}%
  
  \begin{subfigure}[b]{0.16\textwidth}
    \centering
    \includegraphics[width=0.99\textwidth]{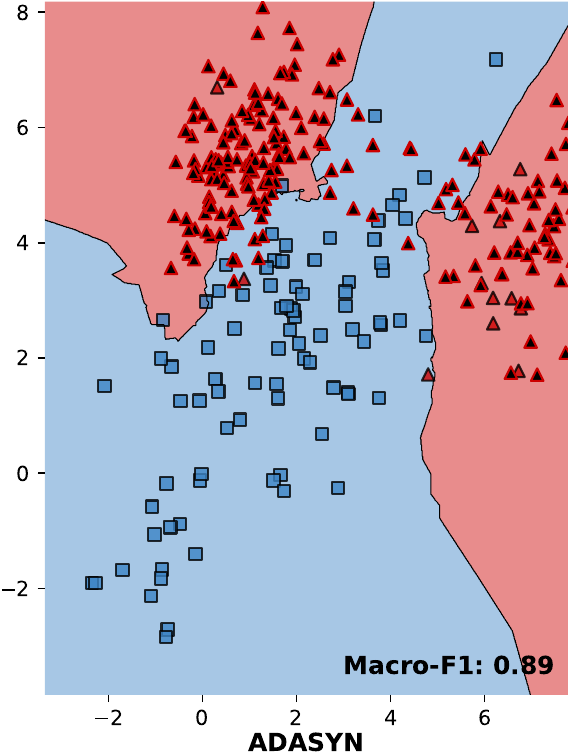}
    \vspace{-10pt}
  \end{subfigure}%
  \begin{subfigure}[b]{0.16\textwidth}
    \centering
    \includegraphics[width=0.99\textwidth]{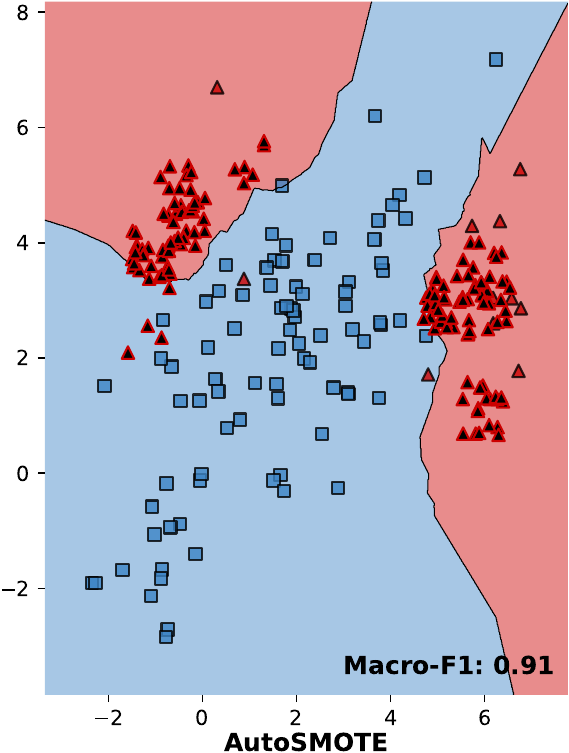}
    \vspace{-10pt}
  \end{subfigure}%
  \begin{subfigure}[b]{0.16\textwidth}
    \centering
    \includegraphics[width=0.99\textwidth]{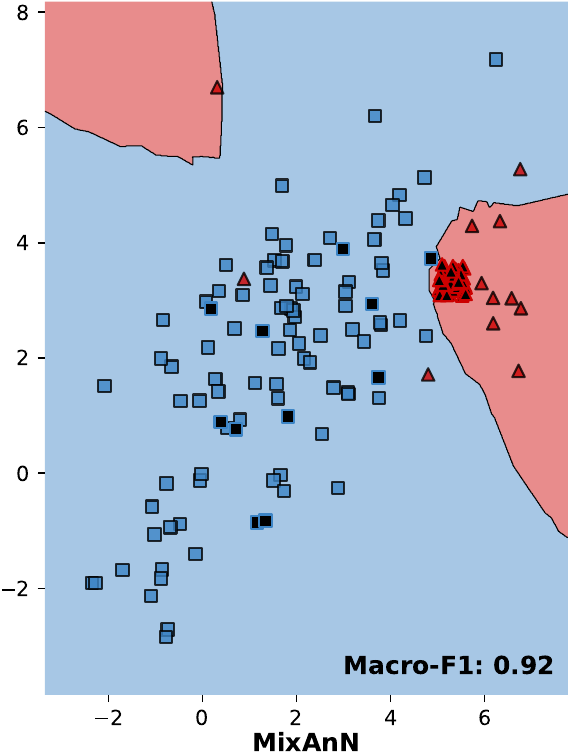}
    \vspace{-10pt}
  \end{subfigure}%
  \vspace{-10pt}
  \caption{Visualization of the generated synthetic samples and the decision boundary of DecisionTree  on a toy data using AutoSMOTE and other over-sampling techniques.}
  \label{fig:case}
  \vspace{-15pt}
\end{figure}

\subsection{Case Study}
To answer the \textbf{RQ5}, we apply MixAnN and the baselines to a publicly
available 2-dimensional toy dataset~\footnote{\url{https://github.com/shubhomoydas/ad_examples/tree/master/ad_examples/datasets/anomaly/toy2/fullsamples}}. We split 60\% of the data for training, 20\% of the data for validation, and 20\% for testing and use Macro-F1 as the performance metric. Figure~\ref{fig:case} illustrates the generated synthetic samples and the decision boundary obtained
by training a KNN classifier on the over-sampled data. We observe that classical SMOTE-based algorithms and Mixboost tend to generate some noisy samples that interleave with the majorities between the anomalies from two different sides, which degrades the performance. AutoSMOTE is capable of generating synthetic samples that are less noisy. However, it has a limitation in cases where there is a true anomaly located at the center. This can lead to the generation of multiple synthetic anomalies that extend the left upper decision boundary, causing misclassification of genuine normal instances. On the other hand, MixAnN is a method that excels in generating synthetic samples of both normal and anomalous instances without introducing any noise. As a result, MixAnN achieves a higher Macro-F1 score compared to AutoSMOTE. This improvement in performance clearly demonstrates the effectiveness of mixing majority and minority instances guided by a learning-based agent.

% however, due to the true anomaly in the center, several synthetic anomalies are generated to extend the left upper decision boundary, resulting in the misclassification of true normalities. In contrast, MixAnN is capable of generating both synthetic normalities and anomalies with zero noises, so it achieves a better Macro-F1 score, which demonstrates the effectiveness of mixing up normalities and anomalies with the guidance of a learning-based agent.

\section{Conclusion and Future Work}
We propose the MixAnN, a strategic data augmentation framework for imbalanced classification with diverse minorities. We propose an iterative mix-up approach to generalize label information and formulate the iterative mix-up into a Markov decision process.  In addition, we further tailor a deep actor-critic framework to tackle the Markov decision process toward an optimal data augmentation policy. We evaluate MixAnN in comparison to state-of-the-art imbalanced classification algorithms and data augmentation approaches and demonstrate the superiority of MixAnN. In the future, we will explore potential approaches to overcome the overconfidence problem of complex classifiers and further improve the scalability of MixAnN with under-sampling methods.

\bibliographystyle{ACM-Reference-Format}
\balance
\bibliography{ref}


\begin{thebibliography}{45}

%%% the Web address.

\ifx \showCODEN    \undefined \def \showCODEN     #1{\unskip}     \fi
\ifx \showDOI      \undefined \def \showDOI       #1{#1}\fi
\ifx \showISBNx    \undefined \def \showISBNx     #1{\unskip}     \fi
\ifx \showISBNxiii \undefined \def \showISBNxiii  #1{\unskip}     \fi
\ifx \showISSN     \undefined \def \showISSN      #1{\unskip}     \fi
\ifx \showLCCN     \undefined \def \showLCCN      #1{\unskip}     \fi
\ifx \shownote     \undefined \def \shownote      #1{#1}          \fi
\ifx \showarticletitle \undefined \def \showarticletitle #1{#1}   \fi
\ifx \showURL      \undefined \def \showURL       {\relax}        \fi
% invisible to TeX
\providecommand\bibfield[2]{#2}
\providecommand\bibinfo[2]{#2}
\providecommand\natexlab[1]{#1}
\providecommand\showeprint[2][]{arXiv:#2}

\bibitem[Al-Hashedi and Magalingam(2021)]%
        {al2021financial}
\bibfield{author}{\bibinfo{person}{Khaled~Gubran Al-Hashedi} {and}
  \bibinfo{person}{Pritheega Magalingam}.} \bibinfo{year}{2021}\natexlab{}.
\newblock \showarticletitle{Financial fraud detection applying data mining
  techniques: a comprehensive review from 2009 to 2019}.
\newblock \bibinfo{journal}{\emph{Computer Science Review}}
  \bibinfo{volume}{40} (\bibinfo{year}{2021}), \bibinfo{pages}{100402}.
\newblock


\bibitem[Bunkhumpornpat et~al\mbox{.}(2012)]%
        {bunkhumpornpat2012dbsmote}
\bibfield{author}{\bibinfo{person}{Chumphol Bunkhumpornpat},
  \bibinfo{person}{Krung Sinapiromsaran}, {and} \bibinfo{person}{Chidchanok
  Lursinsap}.} \bibinfo{year}{2012}\natexlab{}.
\newblock \showarticletitle{DBSMOTE: density-based synthetic minority
  over-sampling technique}.
\newblock \bibinfo{journal}{\emph{Applied Intelligence}} \bibinfo{volume}{36},
  \bibinfo{number}{3} (\bibinfo{year}{2012}), \bibinfo{pages}{664--684}.
\newblock


\bibitem[Cao et~al\mbox{.}(2019)]%
        {cao2019learning}
\bibfield{author}{\bibinfo{person}{Zhangjie Cao}, \bibinfo{person}{Kaichao
  You}, \bibinfo{person}{Mingsheng Long}, \bibinfo{person}{Jianmin Wang}, {and}
  \bibinfo{person}{Qiang Yang}.} \bibinfo{year}{2019}\natexlab{}.
\newblock \showarticletitle{Learning to transfer examples for partial domain
  adaptation}. In \bibinfo{booktitle}{\emph{Proceedings of the IEEE/CVF
  Conference on Computer Vision and Pattern Recognition}}.
\newblock


\bibitem[Chapelle et~al\mbox{.}(2009)]%
        {chapelle2009semi}
\bibfield{author}{\bibinfo{person}{Olivier Chapelle}, \bibinfo{person}{Bernhard
  Scholkopf}, {and} \bibinfo{person}{Alexander Zien}.}
  \bibinfo{year}{2009}\natexlab{}.
\newblock \showarticletitle{Semi-supervised learning (chapelle, o. et al.,
  eds.; 2006)[book reviews]}.
\newblock \bibinfo{journal}{\emph{IEEE Transactions on Neural Networks}}
  \bibinfo{volume}{20}, \bibinfo{number}{3} (\bibinfo{year}{2009}),
  \bibinfo{pages}{542--542}.
\newblock


\bibitem[Chawla et~al\mbox{.}(2002)]%
        {chawla2002smote}
\bibfield{author}{\bibinfo{person}{Nitesh~V Chawla}, \bibinfo{person}{Kevin~W
  Bowyer}, \bibinfo{person}{Lawrence~O Hall}, {and} \bibinfo{person}{W~Philip
  Kegelmeyer}.} \bibinfo{year}{2002}\natexlab{}.
\newblock \showarticletitle{SMOTE: synthetic minority over-sampling technique}.
\newblock \bibinfo{journal}{\emph{Journal of artificial intelligence research}}
   \bibinfo{volume}{16} (\bibinfo{year}{2002}), \bibinfo{pages}{321--357}.
\newblock


\bibitem[Chen et~al\mbox{.}(2020)]%
        {chen2020measuring}
\bibfield{author}{\bibinfo{person}{Deli Chen}, \bibinfo{person}{Yankai Lin},
  \bibinfo{person}{Wei Li}, \bibinfo{person}{Peng Li}, \bibinfo{person}{Jie
  Zhou}, {and} \bibinfo{person}{Xu Sun}.} \bibinfo{year}{2020}\natexlab{}.
\newblock \showarticletitle{Measuring and relieving the over-smoothing problem
  for graph neural networks from the topological view}. In
  \bibinfo{booktitle}{\emph{Proceedings of the AAAI conference on artificial
  intelligence}}.
\newblock


\bibitem[Chen et~al\mbox{.}(2022a)]%
        {chen2022graph}
\bibfield{author}{\bibinfo{person}{Huiyuan Chen},
  \bibinfo{person}{Chin-Chia~Michael Yeh}, \bibinfo{person}{Fei Wang}, {and}
  \bibinfo{person}{Hao Yang}.} \bibinfo{year}{2022}\natexlab{a}.
\newblock \showarticletitle{Graph neural transport networks with non-local
  attentions for recommender systems}. In \bibinfo{booktitle}{\emph{Proceedings
  of the ACM Web Conference 2022}}. \bibinfo{pages}{1955--1964}.
\newblock


\bibitem[Chen et~al\mbox{.}(2022b)]%
        {chen2022adversarial}
\bibfield{author}{\bibinfo{person}{Huiyuan Chen}, \bibinfo{person}{Kaixiong
  Zhou}, \bibinfo{person}{Kwei-Herng Lai}, \bibinfo{person}{Xia Hu},
  \bibinfo{person}{Fei Wang}, {and} \bibinfo{person}{Hao Yang}.}
  \bibinfo{year}{2022}\natexlab{b}.
\newblock \showarticletitle{Adversarial graph perturbations for recommendations
  at scale}. In \bibinfo{booktitle}{\emph{Proceedings of the 45th International
  ACM SIGIR Conference on Research and Development in Information Retrieval}}.
  \bibinfo{pages}{1854--1858}.
\newblock


\bibitem[Dabouei et~al\mbox{.}(2021)]%
        {dabouei2021supermix}
\bibfield{author}{\bibinfo{person}{Ali Dabouei}, \bibinfo{person}{Sobhan
  Soleymani}, \bibinfo{person}{Fariborz Taherkhani}, {and}
  \bibinfo{person}{Nasser~M Nasrabadi}.} \bibinfo{year}{2021}\natexlab{}.
\newblock \showarticletitle{Supermix: Supervising the mixing data
  augmentation}. In \bibinfo{booktitle}{\emph{Proceedings of the IEEE/CVF
  Conference on Computer Vision and Pattern Recognition}}.
  \bibinfo{pages}{13794--13803}.
\newblock


\bibitem[Haarnoja et~al\mbox{.}(2018)]%
        {haarnoja2018soft}
\bibfield{author}{\bibinfo{person}{Tuomas Haarnoja}, \bibinfo{person}{Aurick
  Zhou}, \bibinfo{person}{Pieter Abbeel}, {and} \bibinfo{person}{Sergey
  Levine}.} \bibinfo{year}{2018}\natexlab{}.
\newblock \showarticletitle{Soft actor-critic: Off-policy maximum entropy deep
  reinforcement learning with a stochastic actor}. In
  \bibinfo{booktitle}{\emph{International conference on machine learning}}.
  \bibinfo{pages}{1861--1870}.
\newblock


\bibitem[Han et~al\mbox{.}(2005)]%
        {han2005borderline}
\bibfield{author}{\bibinfo{person}{Hui Han}, \bibinfo{person}{Wen-Yuan Wang},
  {and} \bibinfo{person}{Bing-Huan Mao}.} \bibinfo{year}{2005}\natexlab{}.
\newblock \showarticletitle{Borderline-SMOTE: a new over-sampling method in
  imbalanced data sets learning}. In \bibinfo{booktitle}{\emph{International
  conference on intelligent computing}}. \bibinfo{pages}{878--887}.
\newblock


\bibitem[Han et~al\mbox{.}(2022a)]%
        {han2022adbench}
\bibfield{author}{\bibinfo{person}{Songqiao Han}, \bibinfo{person}{Xiyang Hu},
  \bibinfo{person}{Hailiang Huang}, \bibinfo{person}{Mingqi Jiang}, {and}
  \bibinfo{person}{Yue Zhao}.} \bibinfo{year}{2022}\natexlab{a}.
\newblock \showarticletitle{ADBench: Anomaly Detection Benchmark}. In
  \bibinfo{booktitle}{\emph{Neural Information Processing Systems}}.
\newblock


\bibitem[Han et~al\mbox{.}(2022b)]%
        {han2022g}
\bibfield{author}{\bibinfo{person}{Xiaotian Han}, \bibinfo{person}{Zhimeng
  Jiang}, \bibinfo{person}{Ninghao Liu}, {and} \bibinfo{person}{Xia Hu}.}
  \bibinfo{year}{2022}\natexlab{b}.
\newblock \showarticletitle{G-Mixup: Graph Data Augmentation for Graph
  Classification}. In \bibinfo{booktitle}{\emph{International Conference on
  Machine Learning}}.
\newblock


\bibitem[He et~al\mbox{.}(2008)]%
        {he2008adasyn}
\bibfield{author}{\bibinfo{person}{Haibo He}, \bibinfo{person}{Yang Bai},
  \bibinfo{person}{Edwardo~A Garcia}, {and} \bibinfo{person}{Shutao Li}.}
  \bibinfo{year}{2008}\natexlab{}.
\newblock \showarticletitle{ADASYN: Adaptive synthetic sampling approach for
  imbalanced learning}. In \bibinfo{booktitle}{\emph{2008 IEEE international
  joint conference on neural networks}}. \bibinfo{pages}{1322--1328}.
\newblock


\bibitem[Hsu and Liu(2021)]%
        {hsu2021multiple}
\bibfield{author}{\bibinfo{person}{Chia-Yu Hsu} {and} \bibinfo{person}{Wei-Chen
  Liu}.} \bibinfo{year}{2021}\natexlab{}.
\newblock \showarticletitle{Multiple time-series convolutional neural network
  for fault detection and diagnosis and empirical study in semiconductor
  manufacturing}.
\newblock \bibinfo{journal}{\emph{Journal of Intelligent Manufacturing}}
  \bibinfo{volume}{32} (\bibinfo{year}{2021}), \bibinfo{pages}{823--836}.
\newblock


\bibitem[Ienco et~al\mbox{.}(2016)]%
        {ienco2016semisupervised}
\bibfield{author}{\bibinfo{person}{Dino Ienco}, \bibinfo{person}{Ruggero~G
  Pensa}, {and} \bibinfo{person}{Rosa Meo}.} \bibinfo{year}{2016}\natexlab{}.
\newblock \showarticletitle{A semisupervised approach to the detection and
  characterization of outliers in categorical data}.
\newblock \bibinfo{journal}{\emph{IEEE transactions on neural networks and
  learning systems}} \bibinfo{volume}{28}, \bibinfo{number}{5}
  (\bibinfo{year}{2016}), \bibinfo{pages}{1017--1029}.
\newblock


\bibitem[Kabra et~al\mbox{.}(2020)]%
        {kabra2020mixboost}
\bibfield{author}{\bibinfo{person}{Anubha Kabra}, \bibinfo{person}{Ayush
  Chopra}, \bibinfo{person}{Nikaash Puri}, \bibinfo{person}{Pinkesh Badjatiya},
  \bibinfo{person}{Sukriti Verma}, \bibinfo{person}{Piyush Gupta},
  {et~al\mbox{.}}} \bibinfo{year}{2020}\natexlab{}.
\newblock \showarticletitle{MixBoost: Synthetic Oversampling with Boosted Mixup
  for Handling Extreme Imbalance}.
\newblock \bibinfo{journal}{\emph{arXiv preprint arXiv:2009.01571}}
  (\bibinfo{year}{2020}).
\newblock


\bibitem[Kim et~al\mbox{.}(2020)]%
        {kim2020ai}
\bibfield{author}{\bibinfo{person}{Aechan Kim}, \bibinfo{person}{Mohyun Park},
  {and} \bibinfo{person}{Dong~Hoon Lee}.} \bibinfo{year}{2020}\natexlab{}.
\newblock \showarticletitle{AI-IDS: Application of deep learning to real-time
  Web intrusion detection}.
\newblock \bibinfo{journal}{\emph{IEEE Access}} (\bibinfo{year}{2020}).
\newblock


\bibitem[Kov{\'a}cs(2019)]%
        {kovacs2019smote}
\bibfield{author}{\bibinfo{person}{Gy{\"o}rgy Kov{\'a}cs}.}
  \bibinfo{year}{2019}\natexlab{}.
\newblock \showarticletitle{Smote-variants: A python implementation of 85
  minority oversampling techniques}.
\newblock \bibinfo{journal}{\emph{Neurocomputing}}  \bibinfo{volume}{366}
  (\bibinfo{year}{2019}), \bibinfo{pages}{352--354}.
\newblock


\bibitem[Lai et~al\mbox{.}(2023)]%
        {lai2023context}
\bibfield{author}{\bibinfo{person}{Kwei-Herng Lai}, \bibinfo{person}{Lan Wang},
  \bibinfo{person}{Huiyuan Chen}, \bibinfo{person}{Kaixiong Zhou},
  \bibinfo{person}{Fei Wang}, \bibinfo{person}{Hao Yang}, {and}
  \bibinfo{person}{Xia Hu}.} \bibinfo{year}{2023}\natexlab{}.
\newblock \showarticletitle{Context-aware Domain Adaptation for Time Series
  Anomaly Detection}. In \bibinfo{booktitle}{\emph{Proceedings of the 2023 SIAM
  International Conference on Data Mining}}. \bibinfo{pages}{676--684}.
\newblock


\bibitem[Li et~al\mbox{.}(2021)]%
        {li2021autobalance}
\bibfield{author}{\bibinfo{person}{Mingchen Li}, \bibinfo{person}{Xuechen
  Zhang}, \bibinfo{person}{Christos Thrampoulidis}, \bibinfo{person}{Jiasi
  Chen}, {and} \bibinfo{person}{Samet Oymak}.} \bibinfo{year}{2021}\natexlab{}.
\newblock \showarticletitle{AutoBalance: Optimized Loss Functions for
  Imbalanced Data}. In \bibinfo{booktitle}{\emph{Advances in Neural Information
  Processing Systems}}.
\newblock


\bibitem[Lillicrap et~al\mbox{.}(2015)]%
        {lillicrap2015continuous}
\bibfield{author}{\bibinfo{person}{Timothy~P Lillicrap},
  \bibinfo{person}{Jonathan~J Hunt}, \bibinfo{person}{Alexander Pritzel},
  \bibinfo{person}{Nicolas Heess}, \bibinfo{person}{Tom Erez},
  \bibinfo{person}{Yuval Tassa}, \bibinfo{person}{David Silver}, {and}
  \bibinfo{person}{Daan Wierstra}.} \bibinfo{year}{2015}\natexlab{}.
\newblock \showarticletitle{Continuous control with deep reinforcement
  learning}.
\newblock \bibinfo{journal}{\emph{arXiv preprint arXiv:1509.02971}}
  (\bibinfo{year}{2015}).
\newblock


\bibitem[Ling et~al\mbox{.}(2023a)]%
        {ling2023graph}
\bibfield{author}{\bibinfo{person}{Hongyi Ling}, \bibinfo{person}{Zhimeng
  Jiang}, \bibinfo{person}{Meng Liu}, \bibinfo{person}{Shuiwang Ji}, {and}
  \bibinfo{person}{Na Zou}.} \bibinfo{year}{2023}\natexlab{a}.
\newblock \showarticletitle{Graph Mixup with Soft Alignments}. In
  \bibinfo{booktitle}{\emph{International Conference on Machine Learning}}.
\newblock


\bibitem[Ling et~al\mbox{.}(2023b)]%
        {ling2023learning}
\bibfield{author}{\bibinfo{person}{Hongyi Ling}, \bibinfo{person}{Zhimeng
  Jiang}, \bibinfo{person}{Youzhi Luo}, \bibinfo{person}{Shuiwang Ji}, {and}
  \bibinfo{person}{Na Zou}.} \bibinfo{year}{2023}\natexlab{b}.
\newblock \showarticletitle{Learning Fair Graph Representations via Automated
  Data Augmentations}. In \bibinfo{booktitle}{\emph{International Conference on
  Learning Representations}}.
\newblock


\bibitem[Liu and Zhou(2006)]%
        {liu2006influence}
\bibfield{author}{\bibinfo{person}{Xu-Ying Liu} {and} \bibinfo{person}{Zhi-Hua
  Zhou}.} \bibinfo{year}{2006}\natexlab{}.
\newblock \showarticletitle{The influence of class imbalance on cost-sensitive
  learning: An empirical study}. In \bibinfo{booktitle}{\emph{International
  Conference on Data Mining}}.
\newblock


\bibitem[Liu et~al\mbox{.}(2020)]%
        {liu2020mesa}
\bibfield{author}{\bibinfo{person}{Zhining Liu}, \bibinfo{person}{Pengfei Wei},
  \bibinfo{person}{Jing Jiang}, \bibinfo{person}{Wei Cao},
  \bibinfo{person}{Jiang Bian}, {and} \bibinfo{person}{Yi Chang}.}
  \bibinfo{year}{2020}\natexlab{}.
\newblock \showarticletitle{MESA: Boost Ensemble Imbalanced Learning with
  MEta-SAmpler}. In \bibinfo{booktitle}{\emph{Conference on Neural Information
  Processing Systems}}.
\newblock


\bibitem[Nguyen et~al\mbox{.}(2015)]%
        {nguyen2015deep}
\bibfield{author}{\bibinfo{person}{Anh Nguyen}, \bibinfo{person}{Jason
  Yosinski}, {and} \bibinfo{person}{Jeff Clune}.}
  \bibinfo{year}{2015}\natexlab{}.
\newblock \showarticletitle{Deep neural networks are easily fooled: High
  confidence predictions for unrecognizable images}. In
  \bibinfo{booktitle}{\emph{Proceedings of the IEEE conference on computer
  vision and pattern recognition}}. \bibinfo{pages}{427--436}.
\newblock


\bibitem[Nguyen et~al\mbox{.}(2011)]%
        {nguyen2011borderline}
\bibfield{author}{\bibinfo{person}{Hien~M Nguyen}, \bibinfo{person}{Eric~W
  Cooper}, {and} \bibinfo{person}{Katsuari Kamei}.}
  \bibinfo{year}{2011}\natexlab{}.
\newblock \showarticletitle{Borderline over-sampling for imbalanced data
  classification}.
\newblock \bibinfo{journal}{\emph{International Journal of Knowledge
  Engineering and Soft Data Paradigms}} \bibinfo{volume}{3},
  \bibinfo{number}{1} (\bibinfo{year}{2011}), \bibinfo{pages}{4--21}.
\newblock


\bibitem[Pang et~al\mbox{.}(2019b)]%
        {pang2019weak}
\bibfield{author}{\bibinfo{person}{Guansong Pang}, \bibinfo{person}{Chunhua
  Shen}, \bibinfo{person}{Huidong Jin}, {and} \bibinfo{person}{Anton van~den
  Hengel}.} \bibinfo{year}{2019}\natexlab{b}.
\newblock \showarticletitle{Deep weakly-supervised anomaly detection}.
\newblock \bibinfo{journal}{\emph{arXiv preprint arXiv:1910.13601}}
  (\bibinfo{year}{2019}).
\newblock


\bibitem[Pang et~al\mbox{.}(2019a)]%
        {pang2019deep}
\bibfield{author}{\bibinfo{person}{Guansong Pang}, \bibinfo{person}{Chunhua
  Shen}, {and} \bibinfo{person}{Anton van~den Hengel}.}
  \bibinfo{year}{2019}\natexlab{a}.
\newblock \showarticletitle{Deep anomaly detection with deviation networks}. In
  \bibinfo{booktitle}{\emph{Proceedings of the 25th ACM SIGKDD international
  conference on knowledge discovery \& data mining}}.
  \bibinfo{pages}{353--362}.
\newblock


\bibitem[Pang et~al\mbox{.}(2021)]%
        {pang2021toward}
\bibfield{author}{\bibinfo{person}{Guansong Pang}, \bibinfo{person}{Anton
  van~den Hengel}, \bibinfo{person}{Chunhua Shen}, {and}
  \bibinfo{person}{Longbing Cao}.} \bibinfo{year}{2021}\natexlab{}.
\newblock \showarticletitle{Toward deep supervised anomaly detection:
  Reinforcement learning from partially labeled anomaly data}. In
  \bibinfo{booktitle}{\emph{Proceedings of the 27th ACM SIGKDD conference on
  knowledge discovery \& data mining}}. \bibinfo{pages}{1298--1308}.
\newblock


\bibitem[Perini et~al\mbox{.}(2020)]%
        {perini2020quantifying}
\bibfield{author}{\bibinfo{person}{Lorenzo Perini}, \bibinfo{person}{Vincent
  Vercruyssen}, {and} \bibinfo{person}{Jesse Davis}.}
  \bibinfo{year}{2020}\natexlab{}.
\newblock \showarticletitle{Quantifying the confidence of anomaly detectors in
  their example-wise predictions}. In \bibinfo{booktitle}{\emph{Joint European
  Conference on Machine Learning and Knowledge Discovery in Databases}}.
  Springer, \bibinfo{pages}{227--243}.
\newblock


\bibitem[Ruff et~al\mbox{.}(2020)]%
        {ruff2020deep}
\bibfield{author}{\bibinfo{person}{Lukas Ruff}, \bibinfo{person}{Robert~A.
  Vandermeulen}, \bibinfo{person}{Nico G{\"o}rnitz}, \bibinfo{person}{Alexander
  Binder}, \bibinfo{person}{Emmanuel M{\"u}ller}, \bibinfo{person}{Klaus-Robert
  M{\"u}ller}, {and} \bibinfo{person}{Marius Kloft}.}
  \bibinfo{year}{2020}\natexlab{}.
\newblock \showarticletitle{Deep semi-supervised anomaly detection}. In
  \bibinfo{booktitle}{\emph{International Conference on Learning
  Representations}}.
\newblock


\bibitem[Shorten and Khoshgoftaar(2019)]%
        {shorten2019survey}
\bibfield{author}{\bibinfo{person}{Connor Shorten} {and}
  \bibinfo{person}{Taghi~M Khoshgoftaar}.} \bibinfo{year}{2019}\natexlab{}.
\newblock \showarticletitle{A survey on image data augmentation for deep
  learning}.
\newblock \bibinfo{journal}{\emph{Journal of big data}} \bibinfo{volume}{6},
  \bibinfo{number}{1} (\bibinfo{year}{2019}), \bibinfo{pages}{1--48}.
\newblock


\bibitem[Siriseriwan and Sinapiromsaran(2017)]%
        {siriseriwan2017adaptive}
\bibfield{author}{\bibinfo{person}{Wacharasak Siriseriwan} {and}
  \bibinfo{person}{Krung Sinapiromsaran}.} \bibinfo{year}{2017}\natexlab{}.
\newblock \showarticletitle{Adaptive neighbor synthetic minority oversampling
  technique under 1NN outcast handling}.
\newblock \bibinfo{journal}{\emph{Songklanakarin J. Sci. Technol}}
  \bibinfo{volume}{39}, \bibinfo{number}{5} (\bibinfo{year}{2017}),
  \bibinfo{pages}{565--576}.
\newblock


\bibitem[Sutton and Barto(2018)]%
        {sutton2018reinforcement}
\bibfield{author}{\bibinfo{person}{Richard~S Sutton} {and}
  \bibinfo{person}{Andrew~G Barto}.} \bibinfo{year}{2018}\natexlab{}.
\newblock \bibinfo{booktitle}{\emph{Reinforcement learning: An introduction}}.
\newblock \bibinfo{publisher}{MIT press}.
\newblock


\bibitem[Wang et~al\mbox{.}(2023)]%
        {song2023}
\bibfield{author}{\bibinfo{person}{Song Wang}, \bibinfo{person}{Xingbo Fu},
  \bibinfo{person}{Kaize Ding}, \bibinfo{person}{Chen Chen},
  \bibinfo{person}{Huiyuan Chen}, {and} \bibinfo{person}{Jundong Li}.}
  \bibinfo{year}{2023}\natexlab{}.
\newblock \showarticletitle{Federated Few-Shot Learning}. In
  \bibinfo{booktitle}{\emph{Proceedings of the 29th ACM SIGKDD Conference on
  Knowledge Discovery and Data Mining}}.
\newblock


\bibitem[Wen et~al\mbox{.}(2020)]%
        {wen2020time}
\bibfield{author}{\bibinfo{person}{Qingsong Wen}, \bibinfo{person}{Liang Sun},
  \bibinfo{person}{Fan Yang}, \bibinfo{person}{Xiaomin Song},
  \bibinfo{person}{Jingkun Gao}, \bibinfo{person}{Xue Wang}, {and}
  \bibinfo{person}{Huan Xu}.} \bibinfo{year}{2020}\natexlab{}.
\newblock \showarticletitle{Time series data augmentation for deep learning: A
  survey}.
\newblock \bibinfo{journal}{\emph{arXiv preprint arXiv:2002.12478}}
  (\bibinfo{year}{2020}).
\newblock


\bibitem[Wu and Ortiz(2021)]%
        {wu2021rlad}
\bibfield{author}{\bibinfo{person}{Tong Wu} {and} \bibinfo{person}{Jorge
  Ortiz}.} \bibinfo{year}{2021}\natexlab{}.
\newblock \showarticletitle{Rlad: Time series anomaly detection through
  reinforcement learning and active learning}.
\newblock \bibinfo{journal}{\emph{arXiv preprint arXiv:2104.00543}}
  (\bibinfo{year}{2021}).
\newblock


\bibitem[Yen and Lee(2006)]%
        {yen2006under}
\bibfield{author}{\bibinfo{person}{Show-Jane Yen} {and}
  \bibinfo{person}{Yue-Shi Lee}.} \bibinfo{year}{2006}\natexlab{}.
\newblock \showarticletitle{Under-sampling approaches for improving prediction
  of the minority class in an imbalanced dataset}.
\newblock In \bibinfo{booktitle}{\emph{Intelligent Control and Automation}}.
  \bibinfo{publisher}{Springer}, \bibinfo{pages}{731--740}.
\newblock


\bibitem[Zha et~al\mbox{.}(2023)]%
        {zha2023data}
\bibfield{author}{\bibinfo{person}{Daochen Zha}, \bibinfo{person}{Zaid~Pervaiz
  Bhat}, \bibinfo{person}{Kwei-Herng Lai}, \bibinfo{person}{Fan Yang},
  \bibinfo{person}{Zhimeng Jiang}, \bibinfo{person}{Shaochen Zhong}, {and}
  \bibinfo{person}{Xia Hu}.} \bibinfo{year}{2023}\natexlab{}.
\newblock \showarticletitle{Data-centric artificial intelligence: A survey}.
\newblock \bibinfo{journal}{\emph{arXiv preprint arXiv:2303.10158}}
  (\bibinfo{year}{2023}).
\newblock


\bibitem[Zha et~al\mbox{.}(2022)]%
        {zha2022towards}
\bibfield{author}{\bibinfo{person}{Daochen Zha}, \bibinfo{person}{Kwei-Herng
  Lai}, \bibinfo{person}{Qiaoyu Tan}, \bibinfo{person}{Sirui Ding},
  \bibinfo{person}{Na Zou}, {and} \bibinfo{person}{Xia~Ben Hu}.}
  \bibinfo{year}{2022}\natexlab{}.
\newblock \showarticletitle{Towards automated imbalanced learning with deep
  hierarchical reinforcement learning}. In
  \bibinfo{booktitle}{\emph{Proceedings of the 31st ACM International
  Conference on Information \& Knowledge Management}}.
\newblock


\bibitem[Zha et~al\mbox{.}(2020)]%
        {zha2020meta}
\bibfield{author}{\bibinfo{person}{Daochen Zha}, \bibinfo{person}{Kwei-Herng
  Lai}, \bibinfo{person}{Mingyang Wan}, {and} \bibinfo{person}{Xia Hu}.}
  \bibinfo{year}{2020}\natexlab{}.
\newblock \showarticletitle{Meta-AAD: Active anomaly detection with deep
  reinforcement learning}. In \bibinfo{booktitle}{\emph{2020 IEEE International
  Conference on Data Mining}}.
\newblock


\bibitem[Zhang et~al\mbox{.}(2018)]%
        {zhang2018mixup}
\bibfield{author}{\bibinfo{person}{Hongyi Zhang}, \bibinfo{person}{Moustapha
  Cisse}, \bibinfo{person}{Yann~N. Dauphin}, {and} \bibinfo{person}{David
  Lopez-Paz}.} \bibinfo{year}{2018}\natexlab{}.
\newblock \showarticletitle{mixup: Beyond Empirical Risk Minimization}. In
  \bibinfo{booktitle}{\emph{International Conference on Learning
  Representations}}.
\newblock


\bibitem[Zhao and Hryniewicki(2018)]%
        {zhao2018xgbod}
\bibfield{author}{\bibinfo{person}{Yue Zhao} {and} \bibinfo{person}{Maciej~K
  Hryniewicki}.} \bibinfo{year}{2018}\natexlab{}.
\newblock \showarticletitle{XGBOD: improving supervised outlier detection with
  unsupervised representation learning}. In \bibinfo{booktitle}{\emph{2018
  International Joint Conference on Neural Networks}}.
\newblock


\end{thebibliography}

% \section{Experiment Settings}

% \subsection{Baselines and Training Details}
% For \textbf{single-domain} baselines, we adopt the code of public available GitHub repository of~\cite{malhotra2016lstm}\footnote{\url{https://github.com/PyLink88/Recurrent-Autoencoder}} for EC-LSTM; AE-MLP follows the implementation of PyOD~\footnote{https://github.com/yzhao062/pyod/}. The neural architectures for both baselines are $256-128-128-256$ with different types of neural units.

% For \textbf{dual-domain} baselines, we follow the framework of public available GitHub repository~\footnote{https://github.com/syorami/DDC-transfer-learning} and modify the underlying neural architecture based on the aforementioned single-domain baselines for anomaly detection.  As for SASA, we adopt the implementation of public available GitHub repository~\footnote{https://github.com/DMIRLAB-Group/SASA} and unify the neural architecture of LSTM units and source domain classifier with our framework.

% All algorithms are trained with epoch $5$, batch size $128$ where learning rates for each algorithm are selected from $\{0.05 0.1 0.15 0.2 0.25\}$. The contamination ratios for all algorithms are selected from $\{0.05 0.1 0.15 0.2 0.25 0.3\}$.

\end{document}